\documentclass[journal]{IEEEtran}
\ifCLASSINFOpdf
  \usepackage[pdftex]{graphicx}
\else
\fi
\usepackage{amssymb}
\usepackage[cmex10]{amsmath}
\usepackage{mathtools}
\newtheorem{thm}{Theorem}
\newtheorem{defn}{Definition}

\newtheorem{lema}{Lemma}

\newtheorem{remk}{Remark}

\usepackage{algorithm}
\usepackage{algorithmic}
\usepackage{cases}
\usepackage{array}
\usepackage[tight,footnotesize]{subfigure}
\usepackage{url}
\begin{document}
\title{Decomposition based Transfer Distance Metric Learning for Image Classification}

%\author{Michael~Shell,~\IEEEmembership{Member,~IEEE,}
%        John~Doe,~\IEEEmembership{Fellow,~OSA,}
%        and~Jane~Doe,~\IEEEmembership{Life~Fellow,~IEEE}% <-this % stops a space
%\thanks{M. Shell is with the Department
%of Electrical and Computer Engineering, Georgia Institute of Technology, Atlanta,
%\thanks{Manuscript received April 19, 2005; revised January 11, 2007.}}
\author{Yong~Luo,
        Tongliang~Liu,
        Dacheng~Tao,~\IEEEmembership{Senior Member,~IEEE,}
        and~Chao~Xu,~\IEEEmembership{Member,~IEEE}% <-this % stops a space
\thanks{Y. Luo and C. Xu are with the Key Laboratory of Machine Perception (Ministry of Education),
School of Electronics Engineering and Computer Science, Peking University, Beijing, 100871, China.}
\thanks{T. Liu and D. Tao are affiliated with the Centre for Quantum Computation \& Intelligent Systems and the Faculty of Engineering \& Information Technology, University of Technology, Sydney, 235 Jones Street, Ultimo, NSW 2007, Australia.}

\thanks{\textcopyright 20XX IEEE. Personal use of this material is permitted. Permission from IEEE must be obtained for all other uses, in any current or future media, including
reprinting/republishing this material for advertising or promotional purposes, creating new collective works, for resale or redistribution to servers or lists, or reuse of any copyrighted component of this work in other works.}

}


%\markboth{Journal of \LaTeX\ Class Files,~Vol.~6, No.~1, January~2007}%
%{Shell \MakeLowercase{\textit{et al.}}: Bare Demo of IEEEtran.cls for Journals}

\markboth{$>$ \normalsize{TIP-11260-2013 R}\footnotesize{evision} \normalsize{1} $<$}%
{Shell \MakeLowercase{\textit{et al.}}: Bare Demo of IEEEtran.cls for Journals}




\maketitle

\begin{abstract}
%\boldmath
Distance metric learning (DML) is a critical factor for image analysis and pattern recognition. To learn a robust distance metric for a target task, we need abundant side information (i.e., the similarity/dissimilarity pairwise constraints over the labeled data), which is usually unavailable in practice due to the high labeling cost. This paper considers the transfer learning setting by exploiting the large quantity of side information from certain related, but different source tasks to help with target metric learning (with only a little side information). The state-of-the-art metric learning algorithms usually fail in this setting because the data distributions of the source task and target task are often quite different. We address this problem by assuming that the target distance metric lies in the space spanned by the eigenvectors of the source metrics (or other randomly generated bases). The target metric is represented as a combination of the ``base metrics'', which are computed using the decomposed components of the source metrics (or simply a set of random bases); we call the proposed method, decomposition based transfer DML (DTDML). In particular, DTDML learns a sparse combination of the ``base metrics'' to construct the target metric by forcing the target metric to be close to an integration of the source metrics. The main advantage of the proposed method compared to existing transfer metric learning approaches is that we directly learn the ``base metric'' coefficients instead of the target metric. To this end, far fewer variables need to be learned. We therefore obtain more reliable solutions given the limited side information and the optimization tends to be faster. Experiments on the popular handwritten image (digit, letter) classification and challenge natural image annotation tasks demonstrate the effectiveness of the proposed method.
\end{abstract}


% \fi
\IEEEpeerreviewmaketitle

\section{Introduction}
\label{sec:Introduction}
\IEEEPARstart{T}{he} performance of computer vision, data mining and multimedia systems is heavily dependent on the distance metric between samples. For example, the simple $k$-nearest neighbor ($k$NN) classifier that uses a proper distance metric can be very competitive, and is sometimes superior to other well designed classifiers in many applications such as face recognition, image annotation, etc. In~\cite{KQ-Weinberger-et-al-NIPS-2005}, the authors learn a distance metric for nearest neighbor classification so that the nearest neighbors tend to belong to the same class and the samples from different classes are separated by a large margin. The $k$NN classifier based on the learned metric was shown to be comparable to the state-of-the-art multiclass support vector machine (SVM) in several applications including face recognition and text categorization. A weighted nearest neighbor model was proposed in~\cite{M-Guillaumin-et-al-ICCV-2009} for image annotation that learned a discriminative distance metric. This model was demonstrated empirically to significantly out-perform the state-of-the-art annotation methods on three challenge datasets. Actually, distance metric learning (DML)is also critical to many other popular algorithms, e.g., $k$-means clustering and kernel machines such as SVM.

It is therefore essential to learn a robust distance metric to reveal the data relationships. To achieve this goal, we need a large amount of side information~\cite{EP-Xing-et-al-NIPS-2002} such as the constraints that indicate whether a pair of samples is similar or not. Real-world applications, e.g. image annotation~\cite{Y-Aytar-and-A-Zisserman-ICCV-2011, N-Sawant-et-al-TIP-2013}, usually have few training samples in the instance space of the target learning task due to the high labeling cost. However, we can easily obtain a large number of labeled samples from the instance spaces of different, but related learning tasks, or from the same instance space with different distribution. Therefore, we can leverage the samples from the related tasks for the target task learning. This is known as transfer learning, and the related tasks are usually called source tasks. This article focuses on utilizing the large quantity of side information in the source tasks to discover a reliable distance metric for the target task.

A number of existing metric learning algorithms~\cite{EP-Xing-et-al-NIPS-2002, NCA-J-Goldberger-et-al-NIPS-2004, KQ-Weinberger-et-al-NIPS-2005, ITML-JV-Davis-et-al-ICML-2007, KQ-Weinberger-and-LK-Saul-ICML-2008, RDML-R-Jin-et-al-NIPS-2009} can be utilized to learn a useful distance metric for the source task with adequate training data. The training criterion is usually to minimize the distance between two samples if they are from the same class, and otherwise maximize their distance. However, directly applying the learned source metric to the target task may not result in good performance because it may be biased to the sample distribution of the source task, while the data distributions between the source task and target task maybe quite different. More sophisticated methods should therefore be developed to tackle the metric learning problem in the transfer scenario.

This paper proposes a decomposition-based method for transfer distance metric learning (DTDML) by assuming that the target metric (distance metric of the target task) lies in the space spanned by the eigenvectors of the source metrics, or other randomly generated bases. The target metric is represented as a combination of ``base metrics'' that are derived from the decomposition of the source metrics, or simply computed using the random bases. In particular, DTDML learns a sparse combination of the ``base metrics'' to construct the target metric by forcing the target metric to be close to an integration of the source metrics. The optimization is performed by alternating between the calculation of the ``base metric'' coefficients and source metric integration weights, and both of the two sub-problems can be solved efficiently.

Recent research on transfer metric learning includes the following. In~\cite{LDML-ZJ-Zha-et-al-IJCAI-2009}, the target metric is learned by minimizing the log-determinant divergence between the source metrics and target metric. Zhang and Yeung~\cite{TML-Y-Zhang-and-DY-Yeung-TIST-2012} proposed to learn the task relationships in transfer metric learning, and therefore, allow modeling of negative and zero transfer. These two methods are also optimized using the alternating strategy. However, in each iteration of their alternating procedures, both rely on direct estimation of the target metric and have a large number of $d^2$ variables to be learned. Here, $d$ is the feature dimensionality, which is usually very high for an image, while in DTDML, the number of variables is only $md$ if we use the eigenvectors of $m$ source metrics to construct the ¡°base metrics¡±, and it is common for $m \ll d$. Therefore, we can obtain more reliable solutions, given the limited side information in the target task, and the optimization tends to be faster because we have far fewer variables to be estimated. We adopt Nesterov's optimal method~\cite{Y-Nesterov-MP-2005} for optimization, so do not require costly semi-definite programming in the learning of the target metric, and have a rapid convergence rate. We performed extensive experiments on two popular handwritten image datasets and the challenge NUS-WIDE~\cite{NUSWIDE-TS-Chua-et-al-CIVR-2009} Web image dataset. The results confirmed the effectiveness and efficiency of DTDML.

The article is organized as follows. We summarize closely related works in Section~\ref{sec:Related_Work}. Section 3 includes the description, formulation, and some theoretical analysis of the proposed DTDML. Extensive experiments are presented in Section 4 and we conclude this paper in Section 5.


%\hfill January 11, 2007

\begin{figure*}[!t]
\centering
\includegraphics[width=1.8\columnwidth]{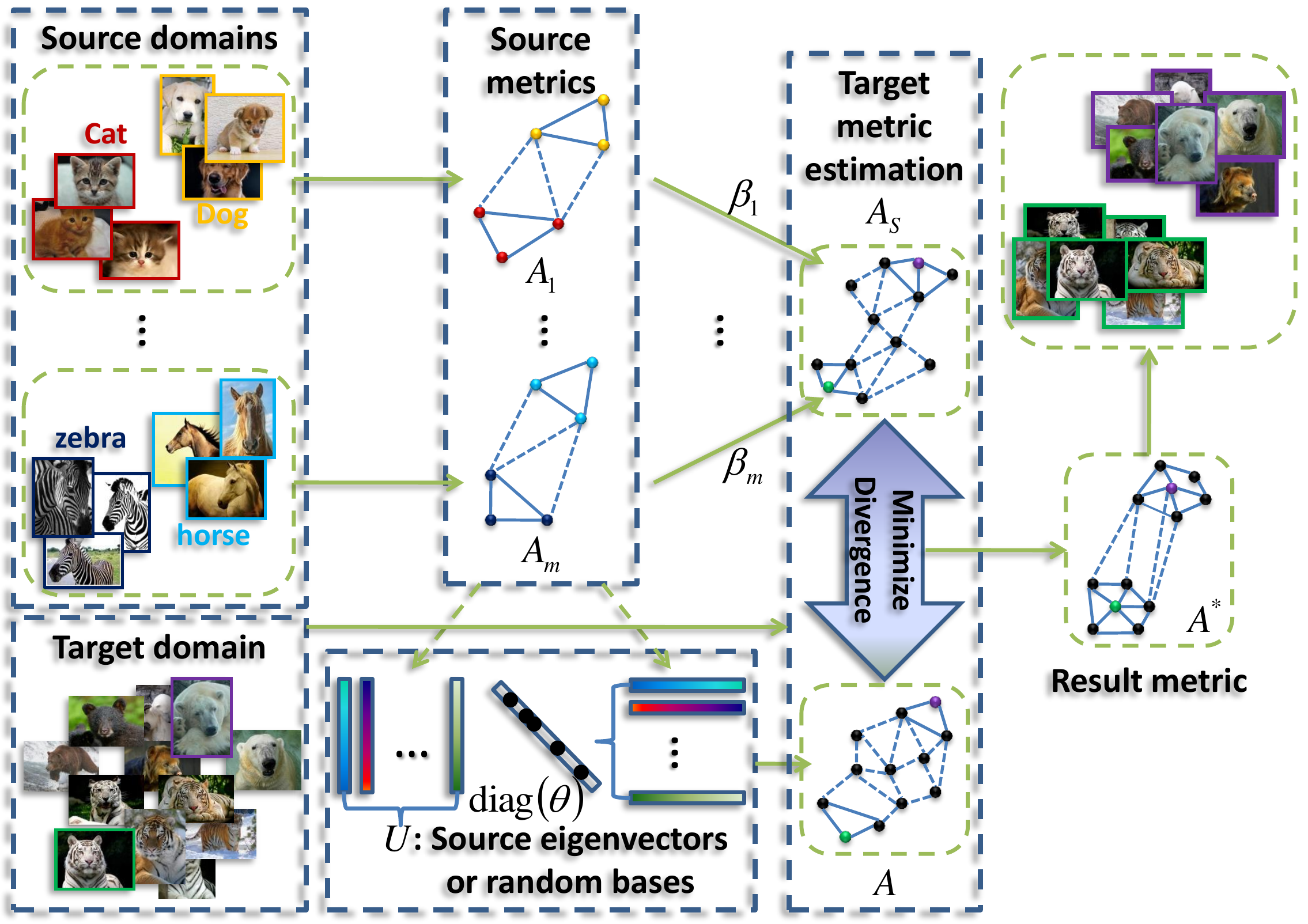}
\caption{Diagram of the proposed DTDML algorithm. Given source domains and a large number of labeled samples for each of them, we learn the corresponding source metrics independently. These metrics are combined as $A_S$ for target metric estimation. At the same time, the source metrics are decomposed into a set of eigenvectors. Alternatively, we can randomly generate a set of bases. The target metric $A = U \mathrm{diag}(\theta) U^T$ is actually a combination of certain ``base metrics'' derived from the source eigenvectors or random bases. By minimizing the divergence between $A_S$ and $A$, we learn the combination coefficients and the source metric integration weights simultaneously, and finally, obtain the result metric.}
\label{fig:System_Diagram}
\end{figure*}

\section{Related Work}
\label{sec:Related_Work}

\subsection{Distance metric learning}
The goal of distance metric learning (DML)~\cite{L-Yang-and-R-Jin-TR-MSU-2006} is to learn an appropriate distance function for a given problem.DML is very important for many learning models, e.g., the kNN rule and SVMs. A popular categorization of the DML methods is: supervised DML~\cite{EP-Xing-et-al-NIPS-2002, M-Schultz-and-T-Joachims-NIPS-2003, NCA-J-Goldberger-et-al-NIPS-2004, KQ-Weinberger-et-al-NIPS-2005, ITML-JV-Davis-et-al-ICML-2007, RDML-R-Jin-et-al-NIPS-2009} and unsupervised DML~\cite{LLE-ST-Roweis-and-LK-Saul-Science-2000, LE-M-Belkin-and-P-Niyogi-NIPS-2001}, according to the underlying learning paradigm. There are also some semi-supervised works that combine these two paradigms~\cite{DCA-SCH-Hoi-et-al-CVPR-2006, MS-Baghshah-and-SB-Shouraki-IJCAI-2009}. Our research is built on supervised metric learning, so we only review some representative works in this category.

A classical algorithm for supervised DML was presented in~\cite{EP-Xing-et-al-NIPS-2002}, where the authors proposed a constrained convex optimization problem for the metric learning. Relevant component analysis (RCA)~\cite{RCA-N-Shental-et-al-ECCV-2002} utilizes the so-called chunklets to learn a metric by reducing the weights of irrelevant dimensions and amplifying the weights of the relevant dimensions. In~\cite{M-Schultz-and-T-Joachims-NIPS-2003}, the relative comparison constraints that can be easily obtained using the query feedbacks were introduced for DML. The formulation is a quadratic programming problem, which was solved by adapting the standard SVM solver. Neighborhood component analysis (NCA)~\cite{NCA-J-Goldberger-et-al-NIPS-2004} learns a metric that directly maximizes the nearest neighbor (NN) classification performance. This is achieved by optimizing the leave-one-out classification error on the training set with stochastic neighborhood selection. Large margin nearest neighbor (LMNN)~\cite{KQ-Weinberger-et-al-NIPS-2005} is also based on NN classification, but using a large margin strategy. From the perspective of information theoretic, Davis et al.~\cite{ITML-JV-Davis-et-al-ICML-2007} proposed to learn a Mahanobis matrix that is close to a given prior distance metric in the sense of differential relative entropy, and simultaneously satisfies the distance constraints s. In \cite{RDML-R-Jin-et-al-NIPS-2009}, an efficient online algorithm was presented for regularized DML, in which it was proved that the generalization error can be independent from the feature dimensionality if appropriate constraints are utilized.

\subsection{Transfer learning}
Transfer learning~\cite{JQ-Candela-et-al-Book-MIT-2009} aims to utilize the knowledge obtained from source domains to help the target domain learning, because the training samples in the target domain are insufficient to train a robust model. Dozens of transfer learning algorithms have been proposed in the literature and can be roughly grouped into homogeneous and heterogeneous transfers. The former refers to samples in target and source domains that are drawn from the same instance space but different distributions~\cite{T-Evgeniou-and-M-Pontil-KDD-2004, JY-Huang-et-al-NIPS-2007, A-Argyriou-et-al-MLJ-2008}, and the latter refers to samples in target and source domains that are drawn from different, but related instance spaces~\cite{J-Blitzer-et-al-EMNLP-2006, A-Argyriou-et-al-NIPS-2007, R-Gupta-and-LA-Ratinov-AAAI-2008}. This research considers the homogeneous setting, and omits are view of the heterogeneous works.

According to~\cite{SJ-Pan-and-Q-Yang-TKDE-2010}, transfer learning can be grouped into instance transfer~\cite{JY-Huang-et-al-NIPS-2007, WY-Dai-et-al-ICML-2007}, feature representation transfer~\cite{A-Argyriou-et-al-MLJ-2008, R-Raina-et-al-ICML-2007}, parameter transfer~\cite{T-Evgeniou-and-M-Pontil-KDD-2004} and relational knowledge transfer~\cite{L-Mihalkova-et-al-AAAI-2007}, based on ``what to transfer''. A kernel mean matching (KMM) method was presented in~\cite{JY-Huang-et-al-NIPS-2007} to match the data distribution of the target domain using the source domain samples. TrAdaboost~\cite{WY-Dai-et-al-ICML-2007} extends AdaBoost to leverage the abundant source data for the target task learning by iteratively filtering out ``bad'' source data. Argyriou et al.~\cite{A-Argyriou-et-al-MLJ-2008} presented a sparse representation based learning algorithm that learns (or selects) some common features shared across related tasks by using a $L_1$-norm regularizer. In~\cite{R-Raina-et-al-ICML-2007}, an unsupervised approach called self-taught learning was proposed to learn features for transfer from unlabeled data. Evgeniou and Pontil~\cite{T-Evgeniou-and-M-Pontil-KDD-2004} learned the parameters of the source and target task simultaneously by assuming the parameter for each task can be separated into two terms, one of which is shared between the source and target task. In~\cite{L-Mihalkova-et-al-AAAI-2007}, the relational knowledge represented with Markov logic networks (MLNs) was transferred from the source domain to the target domain by first constructing a predicate mapping, and then refining the mapped structure in the target domain. There are lots of other works on homogeneous and heterogeneous transfer learning, and we refer to~\cite{SJ-Pan-and-Q-Yang-TKDE-2010} for a more comprehensive survey.

Despite the proposal of many transfer learning algorithms, to the best of our knowledge, only two~\cite{LDML-ZJ-Zha-et-al-IJCAI-2009, TML-Y-Zhang-and-DY-Yeung-TIST-2012} consider homogeneous distance metric transfer. Zha et al.~\cite{LDML-ZJ-Zha-et-al-IJCAI-2009} developed two algorithms for learning a distance metric from a small number of training samples by transferring the prior knowledge from auxiliary data and using a large number of unlabeled samples. Zhang and Yeung~\cite{TML-Y-Zhang-and-DY-Yeung-TIST-2012} proposed a convex formulation for transferring the metric by encoding task relationships in a task covariance matrix. This matrix models positive, negative and zero task correlations. Both algorithms perform well on some applications, but the proposed DTDML will outperform them due the reasons discussed above in Section~\ref{sec:Introduction}. Before presenting the proposed DTDML, we first present certain notations that are used throughout this paper.

\textbf{Notations:} Let $\mathcal{D}=\{(x_i, x_j, y_{ij})\}_{i,j=1}^N$ denotes the training set for the target task, wherein $x_i, x_j \in \mathbb{R}^d$ are vectors of $d$ dimension and $y_{ij}=\pm 1$ indicates $x_i$ and $x_j$ are similar/dissimilar to each other. The number of target training samples $N$ is very small, so we are also given $m$ relevant source training sets, $\mathcal{D}_p = \{ (x_{pi}, x_{pj}, y_{pij}) \}_{i,j=1}^{N_p}, p=1, \ldots , m$, each contains a large amount of training data. In the homogeneous transfer setting, $x_{pi}, x_{pj} \in \mathbb{R}^d$ belong to the same feature space as $x_i, x_j$.

\section{Decomposition based transfer distance metric learning}
\label{sec:DTDML}
Similar to~\cite{TML-Y-Zhang-and-DY-Yeung-TIST-2012}, our method is also built on the regularized DML (RDML)~\cite{RDML-R-Jin-et-al-NIPS-2009} and we introduce it here. In DML, we intend to learn a distance function $dst(x_i, x_j | A)$ parameterized by a distance metric $A$ so that the similarity/dissimilarity between a new instance pair $x_i$ and $x_j$ is reflected by comparing $dst(x_i, x_j | A)$ with a constant threshold $c$. In particular, the regularized distance metric learning (RDML) needs to learn a metric $A$ by the use of the following optimization problem:
\begin{equation}
\label{eq:RDML_Formulation}
\begin{split}
\mathop{\mathrm{argmin}}_A & \ \frac{2}{N(N-1)} \sum_{i<j} g\left( y_{ij} [1-\|x_i - x_j\|_A^2] \right) + \frac{\eta}{2} \|A\|_F^2, \\
\mathrm{s.t.} \ \ & A \succcurlyeq 0.
\end{split}
\end{equation}
where $\|x_i - x_j\|_A^2 = (x_i - x_j)^T A (x_i - x_j)$ is the distance between two samples $x_i$ and $x_j$, and $g(z) = \mathrm{max}(0, b-z)$ is the hinge loss, where $b$ is set to zero in~\cite{RDML-R-Jin-et-al-NIPS-2009}; The Frobenius norm of the metric A, i.e., $\|A\|_F$ is a regularizer that is used to control the model complexity, and $\eta$ is a trade-off parameter. The constraint means that $A$ is positive semi-definite.

An online method was presented in~\cite{RDML-R-Jin-et-al-NIPS-2009} to solve problem (\ref{eq:RDML_Formulation}). However, when training data are limited, RDML performs poorly. Our decomposition based transfer distance metric learning (DTDML) method improves RDML by using training data from certain relevant source domains. As we know that, any metric $A$ can be decomposed as $A = U \Lambda U^T = \sum_{i=1}^d \lambda_i u_i u_i^T$. This indicates that the optimal target metric can be represented as a linear combination of at most $d$ target ``base metrics'' $B_i = u_i u_i^T$. However, the target base metrics are not available. We thus propose to approximate the target base metric by combining some base metrics derived from the source metrics. This approximation is reasonable since the source tasks are related to the target task. Actually, we can also approximate the target base metric using some randomly generated bases and the effectiveness will be demonstrated empirically in our experiments. The proposed target metric learning strategy is advantageous compared to the traditional transfer metric learning algorithm, since we have fewer variables to be learned and thus can obtain more reliable solutions.

The diagram of the proposed DTDML is shown in Fig.~\ref{fig:System_Diagram}. Given $m$ source domains with adequate labeled training data for each, we learn their corresponding metrics $A_p \in \mathbb{R}^{d \times d}, p = 1, \ldots, m$ independently. These source metrics are weighted and integrated as $A_S = \sum_{p=1}^m \alpha_p A_p$, which is used for the target metric estimation later. At the same time, we apply singular value decomposition (SVD) to the obtained source metrics $A_1, \ldots, A_m$ and obtain a set of source eigenvectors $U = [u_1, \ldots, u_n]$, with each $u_r \in \mathbb{R}^d, r = 1, \ldots, n$. Alternatively, $U$ can be a set of randomly generated base vectors. We represent the target metric as $A = U \mathrm{diag}(\theta) U^T = \sum_{r=1}^{n=m \times d} \theta_r u_r u_r^T$, which is actually a combination of base metrics $B_r = u_r u_r^T$. Finally, we learn the source metric integration weights $\alpha$ and the base metric combination coefficients $\theta$ simultaneously by minimizing the divergence between $A_S$ and $A$, as well as leveraging the limited labeled training samples in the target domain. The result metric is given by $A = U \mathrm{diag}(\theta^*) U^T$, where $\theta^*$ is the learned coefficients. The technical details are given below.

\subsection{Problem formulation}
The general formulation of the proposed DTDML for learning the target metric matrix $A$ is given by
\begin{equation}
\label{eq:DTDML_General_Formulation}
\begin{split}
\mathop{\mathrm{argmin}}_{\alpha,\theta} & \ \frac{2}{N(N-1)} \sum_{i<j} V(x_i, x_j, y_{ij}) + \frac{\gamma_A}{2} \|A - A_S\|_F^2 \\
& + \frac{\gamma_B}{2} \|\alpha\|_2^2 + \gamma_C \|\theta\|_1, \\
\mathrm{s.t.} \ \ & \sum_{p=1}^m \alpha_p = 1, \alpha_p \geq 0, p = 1, \ldots, m.
\end{split}
\end{equation}
where $A = \sum_{r=1}^n \theta_r u_r u_r^T$, and the integrated metric $A_S = \sum_{p=1}^m \alpha_p A_p$. The term $\|A - A_S\|_F^2$ is a measure of the difference between $A$ and $A_S$, which are expected to be close. Both $\|\alpha\|_2^2$ and $\|\theta\|_1$ are used to control the model complexity. As depicted above, at most $d$ optimal base metrics are needed to construct the optimal target metric. In practice, most base metric combination coefficients $\lambda_i$ are small and approximate to zero. Therefore, many input base metrics of the proposed model are redundant or noisy. We thus constraint the base metric coefficients $\theta$ to be sparse in order to suppress noisy~\cite{H-Nguyen-et-al-TIP-2013}; $\gamma_A$, $\gamma_B$ and $\gamma_C$ are positive trade-off parameters.

Following~\cite{RDML-R-Jin-et-al-NIPS-2009}, we choose $V(x_i,x_j,y_{ij}) = g(y_{ij} [1 - \|x_i-x_j\|_A^2])$ and adopt the hinge loss~\cite{A-Rodriguez-et-al-TIP-2013} for $g$, i.e., $g(z)=\mathrm{max}(0,b-z)$. Here, $b$ is set to be zero. Then we find the following optimization problem:
\begin{equation}
\label{eq:DTDML_Detailed_Formulation}
\begin{split}
\mathop{\mathrm{argmin}}_{\alpha,\theta} & \ \frac{2}{N(N-1)} \sum_{i<j} g\left( y_{ij} [1 - \|x_i-x_j\|_A^2] \right) \\
& + \frac{\gamma_A}{2} \|A - A_S\|_F^2 + \frac{\gamma_B}{2} \|\alpha\|_2^2 + \gamma_C \|\theta\|_1, \\
\mathrm{s.t.} \ \ & \sum_{p=1}^m \alpha_p = 1, \alpha_p \geq 0, p = 1, \ldots, m.
\end{split}
\end{equation}
For notation simplicity, we denote $x_i$, $x_j$ and $y_{ij}$ as $x_k^1$, $x_k^2$ and $y_k$ respectively, where $k=1, \ldots, N' = \frac{N(N-1)}{2}$. We also set $\delta_k = x_k^1 - x_k^2$ so that $\|x_k^1 - x_k^2\|_A^2 = \sum_{r=1}^n \theta_r \delta_k^T u_r u_r^T \delta_k = \theta^T h_k$ where $h_k = [h_k^1, \ldots, h_k^n]^T$ with each $h_k^r = \delta_k^T u_r u_r^T \delta_k$. Then, the problem (\ref{eq:DTDML_Detailed_Formulation}) becomes
\begin{equation}
\label{eq:DTDML_Detailed_Reformulation}
\begin{split}
\mathop{\mathrm{argmin}}_{\alpha,\theta} & \ \frac{1}{N'} \sum_{k=1}^{N'} g\left( y_k (1 - \theta^T h_k) \right)
+ \frac{\gamma_A}{2} \|A - A_S\|_F^2 \\
& + \frac{\gamma_B}{2} \|\alpha\|_2^2 + \gamma_C \|\theta\|_1, \\
\mathrm{s.t.} \ \ & \sum_{p=1}^m \alpha_p = 1, \alpha_p \geq 0, p = 1, \ldots, m.
\end{split}
\end{equation}
The solution can be obtained by alternating between two sub-problems (which correspond to the minimization w.r.t. $\alpha = [\alpha_1, \ldots, \alpha_m]^T$ and $\theta = [\theta_1, \ldots, \theta_n]^T$ respectively) until convergence.

\subsection{Optimization procedure}
For fixed $\alpha$, the optimization problem with respect to $\theta$ is formulated as
\begin{equation}
\label{eq:DTDML_wrt_Theta}
\mathop{\mathrm{argmin}}_{\theta} F(\theta) = \Phi(\theta) + \Omega(\theta),
\end{equation}
where $\Phi(\theta) = \frac{1}{N'} \sum_{k=1}^{N'} g( y_k (1 - \theta^T h_k) ) + \gamma_C \|\theta\|_1$, and $\Omega(\theta) = \frac{\gamma_A}{2} \|A - A_S\|_F^2$. The loss function $\Phi(\theta)$ is non-differentiable. Hence, we firstly smooth the loss and then use Nesterov's optimal gradient method \cite{Y-Nesterov-MP-2005} to solve (\ref{eq:DTDML_wrt_Theta}). According to~\cite{Y-Nesterov-MP-2005}, the smoothed version of the hinge loss $g(h_k, y_k, \theta) = \mathrm{max}\{0, -y_k(1 - \theta^T h_k)\}$ can be given by
\begin{equation}
\label{eq:Smoothed_g}
g_{\sigma} = \mathop{\mathrm{max}}_{v \in \mathcal{Q}} v_k \left( -y_k (1 - \theta^T h_k) \right) - \frac{\sigma}{2} \|h_k\|_\infty v_k^2,
\end{equation}
where $\mathcal{Q} = \{v: 0 \leq v_k \leq 1, v \in \mathbb{R}^{N'} \}$ and $\sigma$ is the smooth parameter. A larger $\sigma$ induces a more smooth approximation with larger approximation error. On the other hand, a small $\sigma$ induces a slow convergence rate, and thus leads to high time complexity. Therefore, the parameter $\sigma$ should neither be too large nor too small, and we empirically set it as 5 in our implementation. The $\|h_k\|_\infty$ used here is served as a normalization, so that the appropriate value for parameter $\sigma$ does not change too much for different $h_k$. We refer to~\cite{TY-Zhou-et-al-ICDM-2010} for a comprehensive study of the smoothed hinge loss. By setting the objective function of (\ref{eq:Smoothed_g}) to become zero and then projecting $v_k$ on $\mathcal{Q}$, we obtain the following solution,
\begin{equation}
\label{eq:Form_v}
v_k = \mathrm{median} \left\{ \frac{-y_k (1 - \theta^T h_k)}{\sigma \|h_k\|_\infty}, 0, 1 \right\}.
\end{equation}
By substituting the solution (\ref{eq:Form_v}) back into (\ref{eq:Smoothed_g}), we have the piece-wise approximation of $g$, i.e.,
\begin{numcases}
{g_\sigma\!=\!}\mathclap{{}}\!\!\!
\begin{split}
\label{eq:PieceWise_g}
0, \ \ \ \ \ \ \ \ \ \ \ \ \ \ \ \ \ & y_k(1\!-\!\theta^T h_k)\!>\!0; \\
-y_k(1\!-\!\theta^T h_k)\!-\!\frac{\sigma}{2} \|h_k\|_\infty, & y_k(1\!-\!\theta^T h_k)\!<\!-\sigma \|h_k\|_\infty; \\
\frac{\left( y_k(1 - \theta^T h_k) \right)^2}{2 \sigma \|h_k\|_\infty}, \ \ \ \ \ \ & \mathrm{otherwise}.
\end{split}
\end{numcases}
We adopt the Nesterov's method to solve the smoothed version of problem (\ref{eq:DTDML_wrt_Theta}) since it can achieve the optimal convergence rate at $ O(1/k^2)$, which indicates a low time complexity~\cite{Y-Nesterov-MP-2005, TY-Zhou-et-al-ICDM-2010}. To utilize Nesterov's method for optimization, we have to compute the gradient of the smoothed hinge loss to determine the descent direction, as well as the Lipschitz constant to determine the step size of each iteration. We summarize the results in the following theorem.
\begin{thm}
\label{thm:Gradient_Lipschitz_g}
The gradient of the smoothed hinge loss $g_\sigma(\theta)$ is
\begin{equation}
\label{eq:Gradient_g}
\frac{\partial g_\sigma(h_k, y_k, \theta)}{\partial \theta} = y_k h_k v_k.
\end{equation}
The sum of the gradient over all the samples is
\begin{equation}
\label{eq:Gradient_g_All}
\frac{\partial g_\sigma(\theta)}{\partial \theta} = \sum_k y_k h_k v_k = H^\Phi Y v,
\end{equation}
where $H^\Phi = [h_1, \ldots, h_{N'}]$ and $Y = \mathrm{diag}(y)$. The Lipschitz constant of $g_\sigma(\theta)$ is
\begin{equation}
\label{eq:Lipschitz_g}
L^g(\theta) = \frac{N'}{\sigma} \mathop{\mathrm{max}}_k \frac{\|h_k h_k^T\|_2}{\|h_k\|_\infty}.
\end{equation}
\end{thm}
We leave the proof in the Appendix.

Similarly, let $l(\theta)=\|\theta\|_1$, so we have the following piece-wise approximation of $l$ with the smooth parameter $\sigma'$:
\begin{numcases}
{l_\sigma'=}{}
\begin{split}
\label{eq:g_PieceWise}
-\theta_r - \frac{\sigma'}{2}, \ & \ \theta_r < -\sigma'; \\
\theta_r - \frac{\sigma'}{2}, \ & \ \theta_r > \sigma'; \\
\theta_r^2 / (2\sigma'), \ & \ \mathrm{otherwise}.
\end{split}
\end{numcases}
The gradient is given by $\partial (\sum_{r=1}^n l_{\sigma'} (\theta_r)) / \partial \theta = v'$ with each $v'_r = \mathrm{median}\{\theta_r / \sigma', -1, 1\}$ and the Lipschitz constant $L^l(\theta) = 1 / \sigma'$.

In addition, the gradient of $\Omega(\theta)$ is given by
\begin{equation}
\label{eq:Gradient_Omega}
\frac{\partial \Omega(\theta)}{\partial \theta} = H^\Omega \theta - h^\Omega,
\end{equation}
where $H_{st}^\Omega = \gamma_A \mathrm{tr}((u_s u_s^T)(u_t u_t^T))$ and $h_r^\Omega = \gamma_A \mathrm{tr}(A_S^T (u_r u_r^T))$.

Therefore, the gradient of the smoothed $F(\theta)$, is
\begin{equation}
\label{eq:Gradient_Smoothed_F}
\frac{\partial F_\sigma(\theta)}{\partial \theta} = \frac{1}{N'} H^\Phi Y v + \gamma_C v' + H^\Omega \theta - h^\Omega,
\end{equation}
and the Lipschitz constant is
\begin{equation}
\label{eq:Lipschitz_Smoothed_F}
L_\sigma = \frac{1}{\sigma} \mathop{\mathrm{max}}_k \frac{\|h_k h_k^T\|_2}{\|h_k\|_\infty} + \frac{\gamma_C}{\sigma'} + \|H^\Omega\|_2.
\end{equation}

Finally, based on the obtained gradient and Lipschitz constant, we apply Nesterov's method to minimize the smoothed primal $F_\sigma(\theta)$. In the $t$'th iteration round, two auxiliary optimizations are constructed and their solutions are used to build the solution of problem (\ref{eq:DTDML_wrt_Theta}). We use $\theta^t$, $y^t$ and $z^t$ to represent the solutions of DTDML w.r.t. $\theta$ and its two auxiliary optimizations at the $t$'th iteration round, respectively. The Lipschitz constant of $F_\sigma(\theta)$ is $L_\sigma$ and the two auxiliary optimizations are,
\begin{equation}
\notag
\begin{split}
& \mathop{\mathrm{min}}_y \langle \nabla F_\sigma(\theta^t), y-\theta^t \rangle + \frac{L_\sigma}{2} \|y-\theta^t\|_2^2, \\
& \mathop{\mathrm{min}}_z \sum_{i=0}^t \frac{i+1}{2} [F_\sigma(\theta^i) + \langle \nabla F_\sigma(\theta^i), z - \theta^i \rangle] + \frac{L_\sigma}{2} \|z - \hat{\theta}\|_2^2.
\end{split}
\end{equation}
where $\hat{\theta}$ is a guessed solution of $\theta$. By directly setting the gradients of the two objective functions in the auxiliary optimizations as zeros, we can obtain $y^t$ and $z^t$, respectively,
\begin{eqnarray}
\label{eq:Form_y_t}
&& y^t = \theta^t - \frac{1}{L_\sigma} \nabla F_\sigma(\theta^t), \\
\label{eq:Form_z_t}
&& z^t = \hat{\theta} - \frac{1}{L_\sigma} \sum_{i=0}^t \frac{i+1}{2} \nabla F_\sigma(\theta^i).
\end{eqnarray}
The solution after the $t$'th iteration round is the weighted sum of $y^t$ and $z^t$, i.e.,
\begin{equation}
\label{eq:Form_theta_t}
\theta^{t+1} = \frac{2}{t+3} z^t + \frac{t+1}{t+3} y^t.
\end{equation}
The stop criterion is $|F_\sigma(\theta^{t+1}) - F_\sigma(\theta^t)| < \epsilon$. The initialization $\theta^0$ and guessed solution $\hat{\theta}$ are set as the zero vectors.

For fixed $\theta$, the optimization problem with respect to $\alpha$ can be formulated as
\begin{equation}
\label{eq:DTDML_wrt_Alpha}
\begin{split}
\mathop{\mathrm{argmin}}_\alpha & \frac{\gamma_A}{2} \|A - \sum_{p=1}^m \alpha_p A_p\|_F^2 + \frac{\gamma_B}{2} \|\alpha\|_2^2, \\
\mathrm{s.t.} & \ \sum_{p=1}^m \alpha_p = 1, \alpha_p \geq 0, p = 1, \ldots, m.
\end{split}
\end{equation}
This is a standard quadratic programming problem and can be rewritten in compact form as
\begin{equation}
\label{eq:DTDML_wrt_Alpha_Compact}
\begin{split}
\mathop{\mathrm{argmin}}_\alpha & \frac{1}{2} \alpha^T H \alpha - \alpha^T h + \frac{\gamma_B}{2} \|\alpha\|_2^2, \\
\mathrm{s.t.} & \ \sum_{p=1}^m \alpha_p = 1, \alpha_p \geq 0, p = 1, \ldots, m.
\end{split}
\end{equation}
where the constant term has been omitted, $h = [h_1, \ldots, h_m]$ with each $h_p = \gamma_A \mathrm{tr}(A^T A_p)$, and $H$ is a symmetric positive semi-definite matrix with the entry $H_{st} = \gamma_A \mathrm{tr}(A_s^T A_t)$. This is a constrained quadratic optimization problem and can be solved efficiently using the coordinate descent algorithm. In each iteration, we select two elements $\alpha_i$ and $\alpha_j$ to update, and leave the others to be fixed. To satisfy the constraint $\sum_{p=1}^m \alpha_p = 1$, we have $\alpha_i^* + \alpha_j^* = \alpha_i + \alpha_j$, where $\alpha_i^*$ and $\alpha_j^*$ are the solutions of the current iteration. In addition, by using the Lagrangian of (\ref{eq:DTDML_wrt_Alpha_Compact}), we obtain the following updating rule:
\begin{numcases}{}
\label{eq:Update_Rule_Theta}
\begin{split}
& \alpha_i^* = \frac{\gamma_B (\alpha_i+\alpha_j) + (h_i-h_j) + \varepsilon_{ij}}{(H_{ii}-H_{ij}-H_{ji}+H_{jj}) + 2\gamma_B}, \\
& \alpha_j^* = \alpha_i + \alpha_j - \alpha_i^*,
\end{split}
\end{numcases}
where $\varepsilon_{ij} = (H_{ii} - H_{ji} - H_{ij} + H_{jj}) \alpha_i - \sum_{k} (H_{ik} - H_{jk}) \alpha_k$. The obtained $\alpha_i^*$ or $\alpha_j^*$ may violate the constraint $\alpha_p \geq 0$, so we set
\begin{numcases}{}\!\!\!
\notag
\begin{split}
&\alpha_i^*\!=\!0, \alpha_j^*\!=\!\alpha_i\!+\!\alpha_j, \mathrm{if}\ \gamma_B (\alpha_i\!+\!\alpha_j) + (h_i\!-\!h_j) + \varepsilon_{ij} \leq 0, \\
&\alpha_j^*\!=\!0, \alpha_i^*\!=\!\alpha_i\!+\!\alpha_j, \mathrm{if}\ \gamma_B (\alpha_i\!+\!\alpha_j) + (h_j\!-\!h_i) + \varepsilon_{ji} \leq 0.
\end{split}
\end{numcases}

\begin{algorithm}[!t]%[htb]
\caption{The optimization procedure of the proposed DTDML algorithm with automatic determination of the regularization parameters $\gamma_B$ and $\gamma_C$. Both $\rho_C$ and $\rho_B$ are empirically set to one.}
\label{alg:DTDML_Learning_Procedure}
\begin{algorithmic}[1]
\renewcommand{\algorithmicrequire}{\textbf{Initialize:}}
\REQUIRE $\alpha^{(0)}$, $\theta^{(0)}$, $\gamma_B^{(0)}$ and $\gamma_C^{(0)}$. Set $t \leftarrow 0$, construct $A^{(0)} = \sum_{r=1}^n \theta_r^{(0)} u_r u_r^T$ and $A_S^{(0)} = \sum_{p=1}^m \alpha_p^{(0)} A_p$.
\renewcommand{\algorithmicrepeat}{\textbf{Iterate}}
\renewcommand{\algorithmicuntil}{\textbf{Until convergence}}
\REPEAT
\STATE{Optimize
\begin{equation}
\notag
\begin{split}
\theta^{(t+1)} \leftarrow \mathop{\mathrm{argmin}}_{\theta}
& \frac{1}{N'} \sum_{k=1}^{N'} g\left( y_k (1 - \theta^T h_k) \right) \\
& + \frac{\gamma_A}{2} \|A - A_S^{(t)}\|_F^2 + \gamma_C^{(t)} \|\theta\|_1
\end{split}
\end{equation} and update $A^{(t+1)} = \sum_{r=1}^n \theta_r^{(t+1)} u_r u_r^T;$}
\STATE{Optimize
\begin{equation}
\notag
\alpha^{(t+1)} \leftarrow \mathop{\mathrm{argmin}}_{\alpha} \frac{\gamma_A}{2} \|A^{(t+1)} - A_S\|_F^2 + \frac{\gamma_B^{(t)}}{2} \|\alpha\|_2^2
\end{equation} and update $A_S^{(t+1)} = \sum_{p=1}^m \alpha_p^{(t+1)} A_p;$}
\STATE{Compute
\begin{equation}
\notag
\begin{split}
\gamma_C^{(t+1)} = & |\rho_C| \Big[ \frac{1}{N'} \sum_{k=1}^{N'} g\left( y_k (1 - (\theta^{(t+1)})^T h_k) \right) \\
& + \frac{\gamma_A}{2} \|A^{(t+1)} - A_S^{(t)}\|_F^2 \Big] / {\|\theta^{(t+1)}\|_1};
\end{split}
\end{equation}}
\STATE{Compute
\begin{equation}
\notag
\gamma_B^{(t+1)} = |\rho_B| \big[ \gamma_A \|A^{(t+1)} - A_S^{(t+1)}\|_F^2 \big] / \|\alpha^{(t+1)}\|_2^2;
\end{equation}}
\STATE{$t \leftarrow t+1.$}
\UNTIL
\end{algorithmic}
\end{algorithm}

\subsection{Automatic determination of the regularization parameters $\gamma_B$ and $\gamma_C$}
In the proposed model (\ref{eq:DTDML_Detailed_Reformulation}), we have three parameters $\gamma_A$, $\gamma_B$ and $\gamma_C$ to determine. Determination of all these parameters is nontrivial due to the limited number of labeled data available in the target task. Therefore, we present an automatic determination algorithm for the regularization parameters $\gamma_B$ and $\gamma_C$. This algorithm is inappropriate for the determination of $\gamma_A$ because the corresponding regularization term $\|A - A_S\|_F^2$ is a coupling of $\alpha$ and $\theta$.

The algorithm is based on the L-curve, which graphically displays the trade-off between approximation error and solution size as the regularization parameter varies~\cite{S-Oraintara-et-al-ICIP-2000, A-Mirzal-arXiv-2012}. The proper regularization parameter value is associated with the corner of the curve, where both solution and approximation error have small norms. Following~\cite{A-Mirzal-arXiv-2012}, we choose a tangency-based method~\cite{S-Oraintara-et-al-ICIP-2000} to find the L-corner since it has a convergence guarantee and the computation is fast. The procedure is shown in Algorithm~\ref{alg:DTDML_Learning_Procedure}, where $\rho_C$ and $\rho_B$ are slopes of the straight line that are tangent to the L-curves, and are set to be one, empirically, in this paper.

The stopping criterion for terminating the algorithm can be the difference of the objective value $\frac{1}{N'} \sum_{k=1}^{N'} g( y_k (1-\theta^T h_k) ) + \frac{\gamma_A}{2} \|A - A_S\|_F^2 + \frac{\gamma_B}{2} \|\alpha\|_2^2 + \gamma_C \|\theta\|_1$ between two consecutive steps. Alternatively, we can stop the iterations when the variation of $\alpha$ and $\theta$ are both smaller than a predefined threshold. Our implementation is based on the difference of the objective value, i.e., if the value $|O_k - O_{k-1}| / |O_k - O_0|$ is smaller than a predefined threshold, then the iteration stops, where $O_k$ is the objective value of the $k$'th iteration step.

\subsection{Theoretical analysis}
The generalization error bound of the proposed DTDML algorithm is now provided. We derive the generalization bound using the uniform stability~\cite{O-Bousquet-and-A-Elisseeff-JMLR-2002}.

\subsubsection{Uniform Stability}
\begin{defn}
\label{defn:Uniform_Stability}
\emph{(Uniform stability~\cite{O-Bousquet-and-A-Elisseeff-JMLR-2002})}. An algorithm has uniform stability $\beta$ with respect to the loss function $l$ if the following holds
\begin{equation}
\label{eq:Uniform_Stability_Defn}
\forall s \in \mathcal{Z}^m, \forall i \in \{1, \ldots, m\}, \|l(h_s,\cdot) - l(h_{s^i},\cdot)\|_\infty \leq \beta,
\end{equation}
where $\mathcal{Z}$ is the sample space, $h_s$ is the hypothesis function returned by the algorithm learning with the set of samples $s$, and $s^i = \{z_1, \ldots, z_{i-1}, z_{i'}, z_{i+1}, \ldots, z_m\}$ denotes a set of samples with the $i$'th element $z_i$ replaced by $z_{i'}$.
\end{defn}

To obtain the uniform stability, we use the Bregman divergence~\cite{M-Mohri-et-al-Book-MIT-2012}. Bregman divergence is defined for any convex and differentiable function $F: \mathcal{H} \rightarrow \mathbb{R}$ as follows (here $\mathcal{H}$ denotes the Hilbert space):
\begin{equation}
\label{eq:Bregman_Divergence}
\forall f, g \in \mathcal{H}, B_F (f \parallel g) = F(f) - F(g) - \mathrm{tr}(\langle f-g, \nabla F(g) \rangle).
\end{equation}
For non-differential loss function, we use the generalized Bregman divergence. The sub-gradient of $F$ at $h$ (see e.g.,~\cite{GA-Watson-LAA-1992}) is defined as
\begin{equation}
\label{eq:Sub_Gradient}
\partial F(h) = \{g \in \mathcal{H}\ |\ \forall h' \in \mathcal{H}, F(h') - F(h) \geq \mathrm{tr}(\langle h'-h, g \rangle) \}.
\end{equation}
Let $\delta F(h)$ be an arbitrary element of $\partial F(h)$. The generalized Bregman divergence to $F$ is then defined as
\begin{equation}
\label{eq:Non_Diff_Bregman_Divergence}
\forall h',h \in \mathcal{H}, B_F(h' \parallel h) = F(h') - F(h) - \mathrm{tr}(\langle h'-h, \delta F(h)\rangle).
\end{equation}
According to the definition of sub-gradient, we have $B_F (h' \parallel h) \geq 0$ and $B_{P+Q} = B_P + B_Q$ for any convex functions $P$ and $Q$. That is, the generalized Bregman divergence is non-negative and additive.

In addition, to derive the uniform stability, we need the following lemma cited from \cite{RDML-R-Jin-et-al-NIPS-2009} (Proposition 2 therein):
\begin{lema}
\label{lema:Metric_Diff_For_Bound}
For any two distance metrics $A$ and $A'$, the following inequality holds for any sample $z_i$ and $z_j$
\begin{equation}
\label{eq:Metric_Diff_For_Bound}
|V(A, z_i, z_j) - V(A', z_i, z_j)| \leq 4LR^2 \|A - A'\|_F.
\end{equation}
\end{lema}

Then we present the uniform stability for our model.

\begin{thm}
\label{thm:Uniform_Stability}
Let $\beta$ be the uniform stability of the developed algorithm for problem (\ref{eq:DTDML_General_Formulation}) and assume $\|x\|_2 \leq R$ for any sample $x$. Then,
\begin{equation}
\label{eq:Uniform_Stability}
\beta \leq \frac{64 L^2 R^4}{\gamma_A N}.
\end{equation}
where $L$ is the Lipschitz constant of the function $g$.
\end{thm}

The detailed proof of Theorem~\ref{thm:Uniform_Stability} can be found in the Appendix. We then derive the generalization bound via the uniform stability.

\subsubsection{Generalization error bound}
Let $\mathcal{N}$ denote the sample set and $V(A,z_i,z_j) = g(y_{ij} [1 - \|x_i-x_j\|_A^2])$. The empirical risk and expected risk can be defined as $R_\mathcal{N}(A) = \frac{2}{N(N-1)} \sum_{i<j} V(A,z_i,z_j)$ and $R(A)=E_{(z_i,z_j)} [V(A,z_i,z_j)]$, respectively. A probabilistic bound on the defect $R(A) - R_\mathcal{N}(A)$ is called the generalization bound.

The bound can be derived by utilizing the obtained uniform stability and the following McDiarmid inequality~\cite{C-McDiarmid-Survey-1989}.
\begin{thm}
\label{thm:MicDiarmid_Inequality}
\emph{(McDiarmid inequality~\cite{C-McDiarmid-Survey-1989}).} Let $z_1, \ldots, z_N \in \mathcal{Z}$ be a set of $N$ independent random variables and assume that there exist $c_1, \ldots, c_N$ such that $f:\{z_i\}_{i=1}^N \mapsto \mathbb{R}$ satisfying
\begin{equation}
\mathop{\mathrm{sup}}_{\mathclap{z_1, \ldots, z_N,z_{i'}}} |f(z_1, \ldots, z_N) - f(z_1, \ldots, z_{i-1}, z_{i'}, z_{i+1}, z_N)| \leq c_i,
\end{equation}
for all $i \in [1,N]$ and any point $z_{i'} \in \mathcal{Z}$. Let $f(\mathcal{N}) = f(z_1, \ldots, z_N)$. Then, for all $\epsilon > 0$, the following holds:
\begin{equation}
\label{eq:MicDiarmid_Inequality}
Pr(f(\mathcal{N})-E[f(\mathcal{N})] \geq \epsilon) \leq \mathrm{exp}(\frac{-2\epsilon^2}{\sum_{i=1}^N c_i^2}).
\end{equation}
\end{thm}

Then we present the generalization bound for our model.

\begin{thm}
\label{thm:Generalization_Error_Bound}
Let $\mathcal{N}$ be a set of $N$ randomly selected samples and $A_\mathcal{N}$ be the distance metric learned by solving (\ref{eq:DTDML_General_Formulation}). With probability at least $1-\delta$, we have
\begin{equation}
\label{eq:Generalization_Error_Bound}
\begin{split}
R(A_\mathcal{N}) - R_\mathcal{N}(A_\mathcal{N}) \leq \frac{128 L^2 R^4}{\gamma_A N} + M \sqrt{\frac{\mathrm{ln}(1/\delta)}{2N}},
\end{split}
\end{equation}
where
\begin{equation}
\notag
M = \frac{128 L^2 R^4 + 4 \gamma_A g_{A_S} + 16 \sqrt{2 \gamma_A} L R^2 \sqrt{g_{A_S} + \gamma_C \|\theta_S\|_1}}{\gamma_A}.
\end{equation}
Here, $\theta_S$ is a solution of $A = A_S$ and $g_{A_S} = \mathrm{sup}_{z_i,z_j} V(A_S,z_i,z_j)$ is the largest loss when the distance metric is $A_S$.
\end{thm}

To prove Theorem~\ref{thm:Generalization_Error_Bound}, we need an additional lemma:
\begin{lema}
\label{lema:Inequalities_For_Bound}
The following two inequalities hold: 1) $\|A_\mathcal{N} - A_S\|_F \leq \sqrt{\left( 2(g_{A_S} + \gamma_C (\|\theta_S\|_1-\|\theta_\mathcal{N}\|_1)) \right) / \gamma_A}$ and 2) $\|A_\mathcal{N'} - A_S\|_F \leq \sqrt{\left( 2(g_{A_S} + \gamma_C (\|\theta_S\|_1-\|\theta_\mathcal{N'}\|_1)) \right) / \gamma_A} \leq \sqrt{\left( 2(g_{A_S} + \gamma_C \|\theta_S\|_1) \right) / \gamma_A}$.
\end{lema}

The detailed proof of Theorem~\ref{thm:Generalization_Error_Bound} for the bound can be found in the Appendix.
\begin{remk}
\label{remk:Generalization_Error_Bound}
In the upper bound of generalization error, $A_S$ and $\theta_S$ are learned from the source data. They are the information that was transferred from the source data to the target data.
\end{remk}

\section{Experimental evaluation}
\label{sec:Experiments}
This section outlines the validation of the effectiveness of the proposed DTDML empirically on two popular handwritten image datasets, and a challenging natural image dataset. The first two datasets are obtained from~\cite{TML-Y-Zhang-and-DY-Yeung-TIST-2012}. Specifically, we compare the following methods: \\
$\bullet$ \textbf{RDML \cite{RDML-R-Jin-et-al-NIPS-2009}:} an online algorithm that has been demonstrated empirically to be effective and quite efficient in learning a distance metric, and can handle high dimensional data. This algorithm serves as a baseline here since it learns only from the target task and leverages nothing from the source tasks. \\
$\bullet$ \textbf{RDML\_AGG:} a simple aggregation strategy, which is to learn the target metric by directly applying RDML on the training set that consists of data from both the source and target tasks. \\
$\bullet$ \textbf{LDML \cite{LDML-ZJ-Zha-et-al-IJCAI-2009}:} a transfer distance metric learning algorithm that is based on~\cite{ITML-JV-Davis-et-al-ICML-2007}, and is formulated as: \\
\begin{equation}
\label{eq:LDML_Formulation}
\begin{split}
\mathop{\mathrm{argmin}}_A & \sum_{p=1}^m \beta_p \left( \mathrm{tr}(A_p^{-1}A) \right) - \log\det A + \gamma_S \mathrm{tr}(SA) \\
& - \gamma_D \mathrm{tr}(DA) + \gamma_B \|\beta\|_2^2, \\
\mathrm{s.t.} & \ A \succcurlyeq 0, \sum_{p=1}^m \beta_p = 1, \beta_p \geq 0, p = 1, 2, \ldots, m,
\end{split}
\end{equation}
where $S$ and $D$ are matrices of the similar and dissimilar constraints. The above formulation contains a semi-definite programming (SDP) problem, and in our re-implementation it is solved using the SDPT3 solver. According to~\cite{LDML-ZJ-Zha-et-al-IJCAI-2009}, the parameters can be set empirically as $\gamma_D = \frac{1}{4} \gamma_S$ and $\gamma_B = \frac{1}{8} \gamma_S$. Therefore, only $\gamma_S$ needs to be tuned. \\
$\bullet$ \textbf{TML \cite{TML-Y-Zhang-and-DY-Yeung-TIST-2012}:} a recently proposed transfer metric learning algorithm. Similar to~\cite{RDML-R-Jin-et-al-NIPS-2009}, an online algorithm is developed to learn the target metric. The task relationship is learned for transfer by solving a second-order cone programming (SOCP) problem using the CVX solver. In addition, the parameters are automatically determined by adopting a Bayesian regularization scheme for the model. \\
$\bullet$ \textbf{DTDML:} the proposed decomposition based transfer distance metric learning. The parameters $\gamma_B$ and $\gamma_C$ are determined automatically and we only need to optimize $\gamma_A$.

\begin{figure}
\centering
\subfigure{\includegraphics[width=0.7\columnwidth]{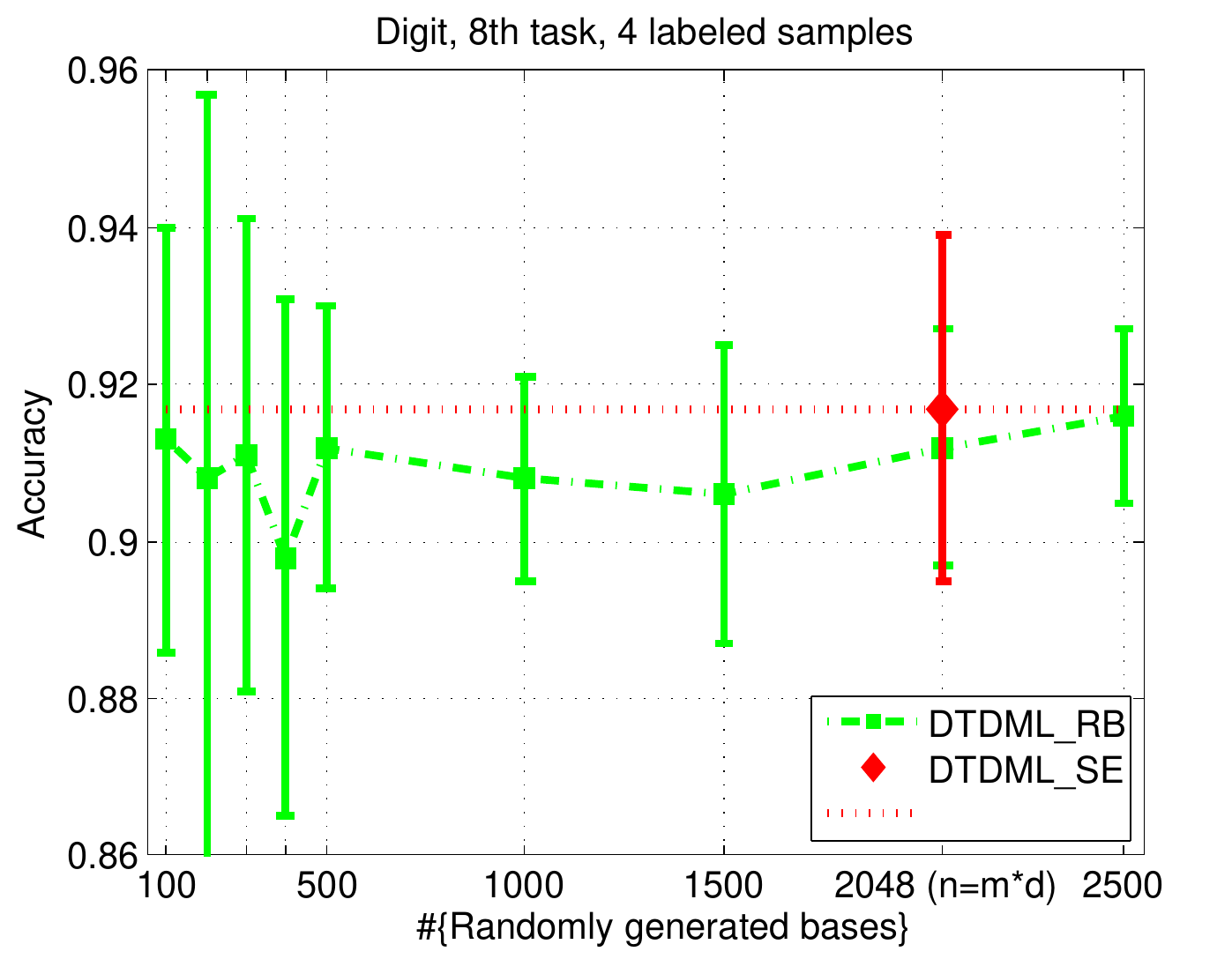}
%\label{subfig:Random_Bases_Digit}
}
\hfil
\subfigure{\includegraphics[width=0.7\columnwidth]{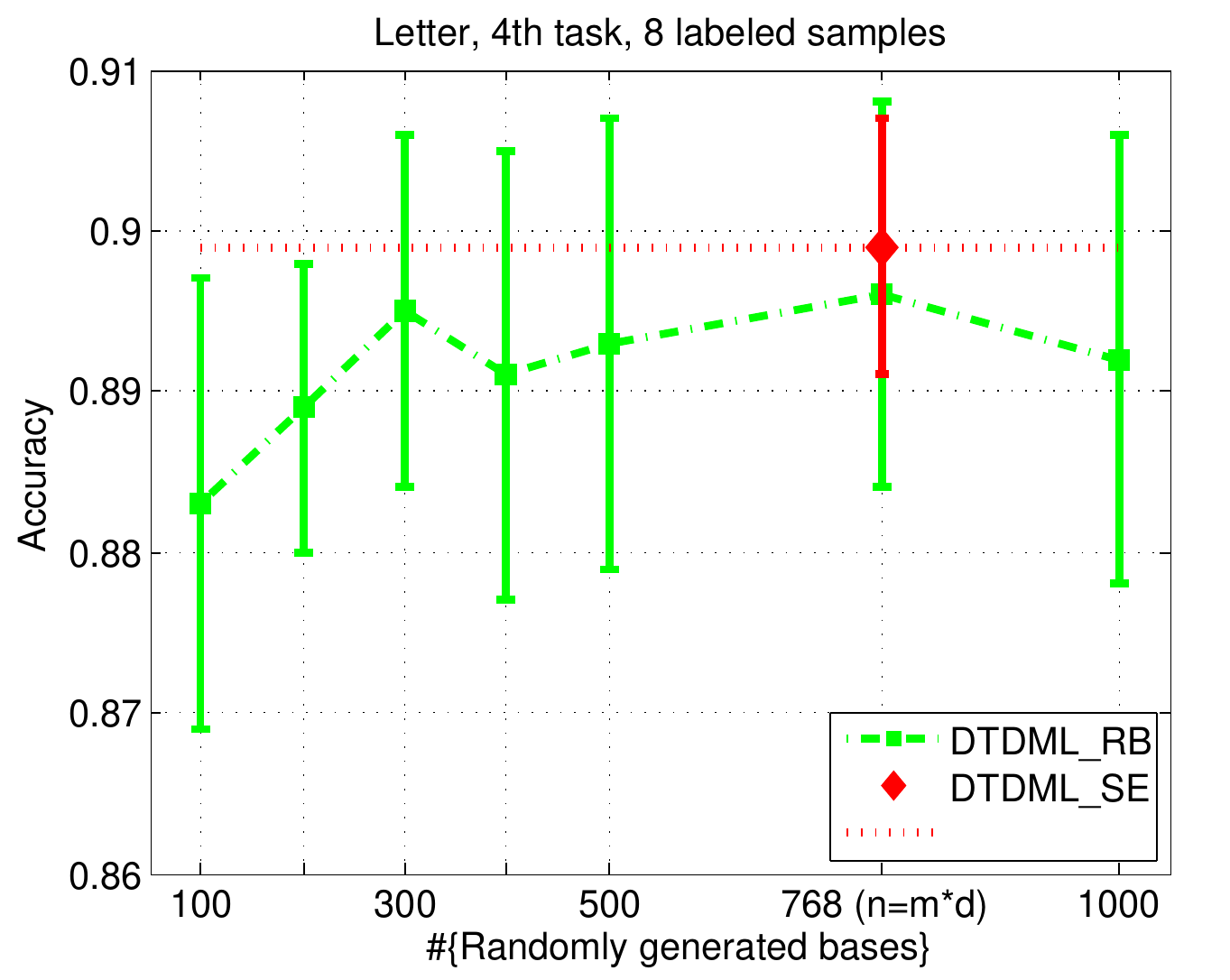}
%\label{subfig:Random_Bases_Letter}
}
\caption{A comparison of DTDML using source eigenvectors (DTDML\_SE) and random bases (DTDML\_RB).A different number of random bases is investigated. (Top row: Handwritten digit; Bottom row: Handwritten letter.)}
\label{fig:Random_Bases}
\end{figure}

We train the source metrics using the RDML method and all the available data in the source tasks. We split the data into equal training and test sets for the target task. The number of labeled samples that are chosen from the training set is gradually increased to see the performance variation w.r.t. the size of the labeled set. We evaluate the learned target metric by applying the 1-nearest-neighbor classifier on the test set. Ten random choices of the labeled samples are used in our experiments. Both the mean and standard deviation of the accuracies are reported.

\subsection{Handwritten image classification}
One of the handwritten image datasets we use is the well-known USPS digit dataset\footnote{\url{http://www.csie.ntu.edu.tw/~cjlin/libsvmtools/datasets/multiclass.html#usps}}, which contains $7,291$ samples. Each sample is an image of size $16 \times 16$ in raw pixels, and the feature dimension $d = 256$. We consider nine classification tasks, i.e., 0/6, 0/8, 1/4, 2/7, 3/5, 4/7, 4/9, 5/8, and 6/8, each corresponding to a classification of two digits. One of the nine tasks is treated as the target task and the others are the source tasks (each task is treated as the target task in turn).

The other is a handwritten letter dataset\footnote{\url{http://ai.stanford.edu/~btaskar/ocr/}}, which is a little different from the dataset presented in~\cite{TML-Y-Zhang-and-DY-Yeung-TIST-2012} since it cannot be downloaded immediately, according to the web link provided. The letter dataset used in this paper consists of $52,152$ samples and the feature dimension is $128$. Six binary classification problems, i.e., c/e, m/n, a/g, a/o, f/t, and h/n, are considered. For each task, we randomly select at most $1,000$ positive and $1,000$ negative samples from the dataset. The experimental settings are the same as for those of the digit classification.

\begin{figure*}
\centering
\subfigure{\includegraphics[width=0.65\columnwidth]{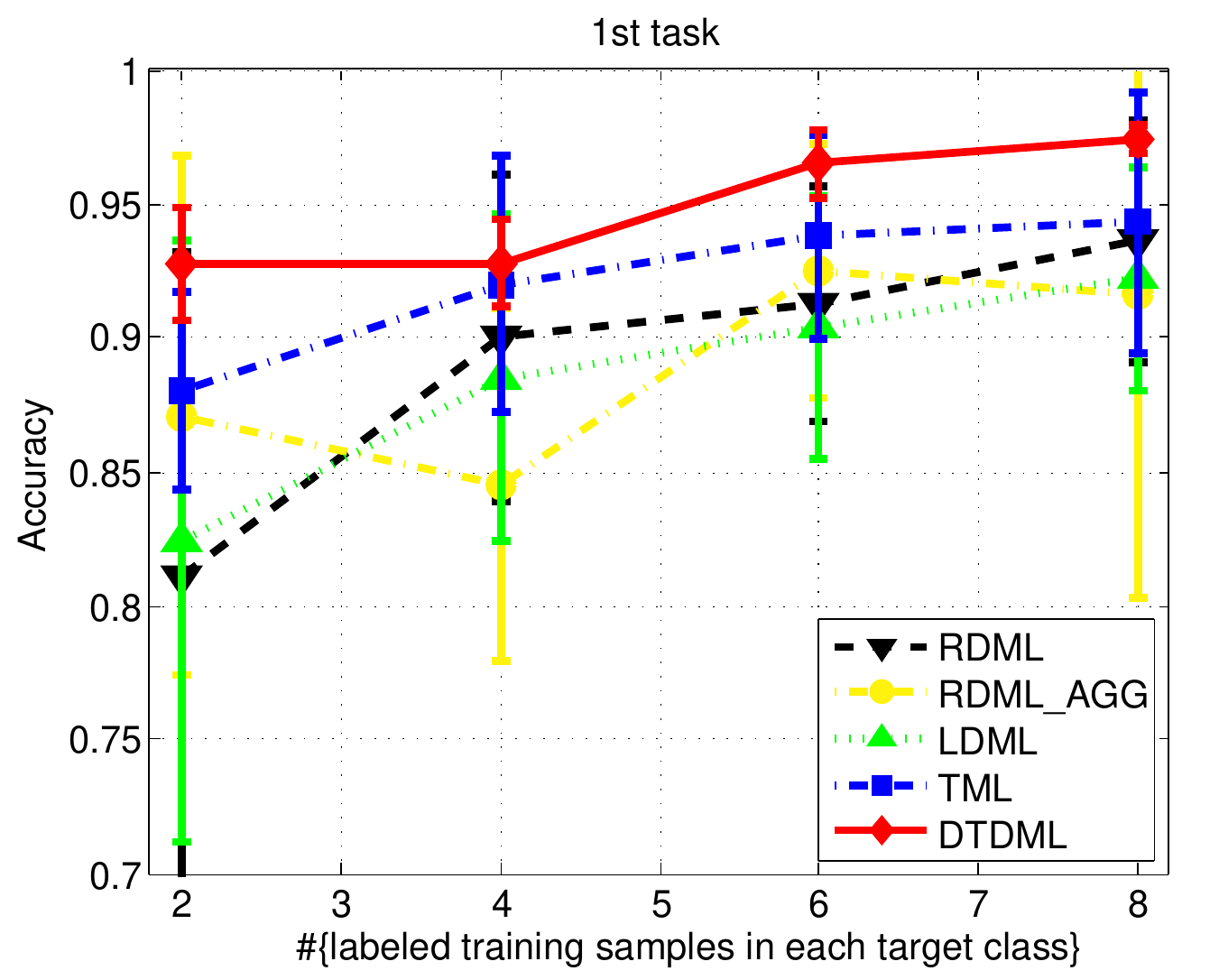}
}
\hfil
\subfigure{\includegraphics[width=0.65\columnwidth]{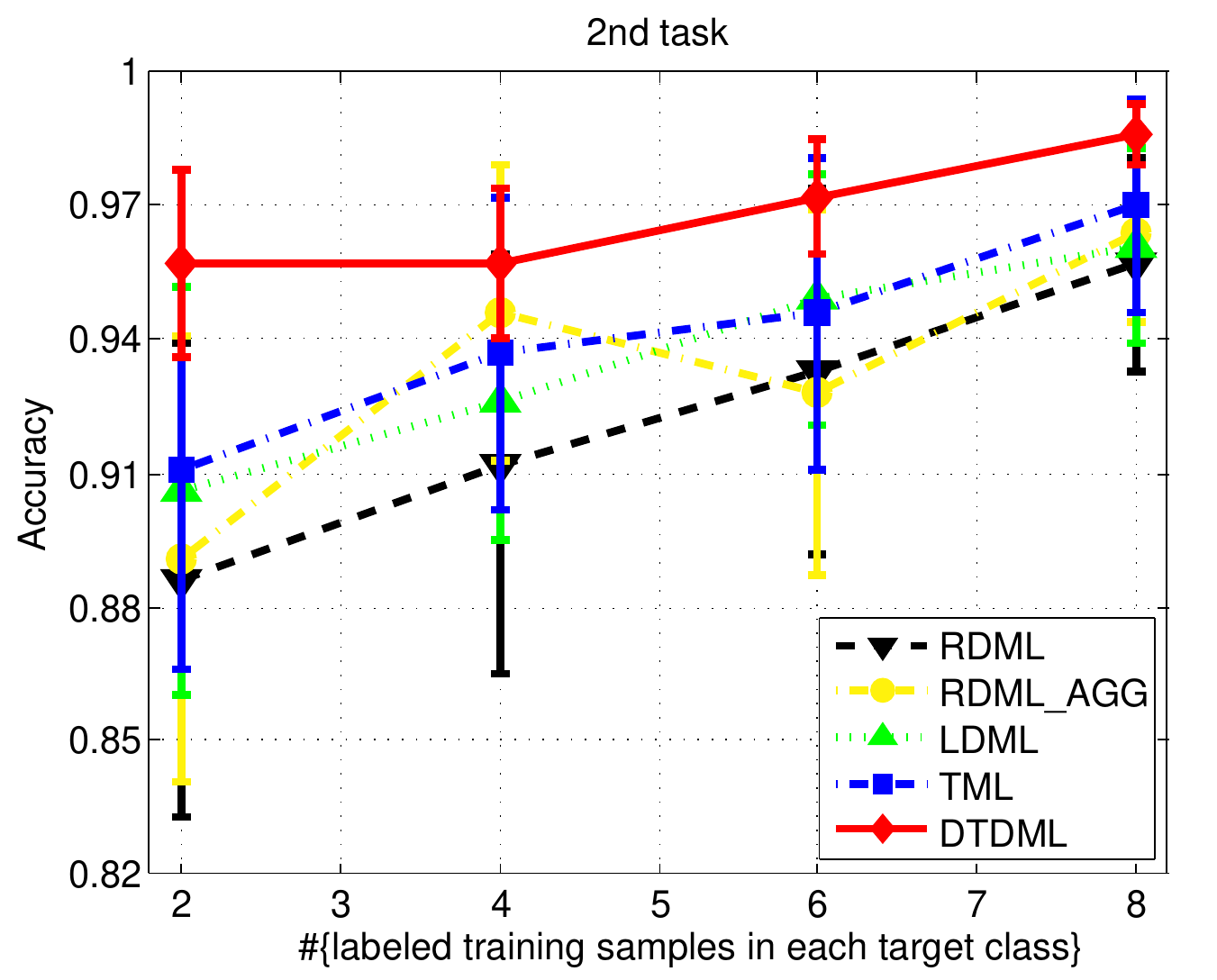}
}
\hfil
\subfigure{\includegraphics[width=0.65\columnwidth]{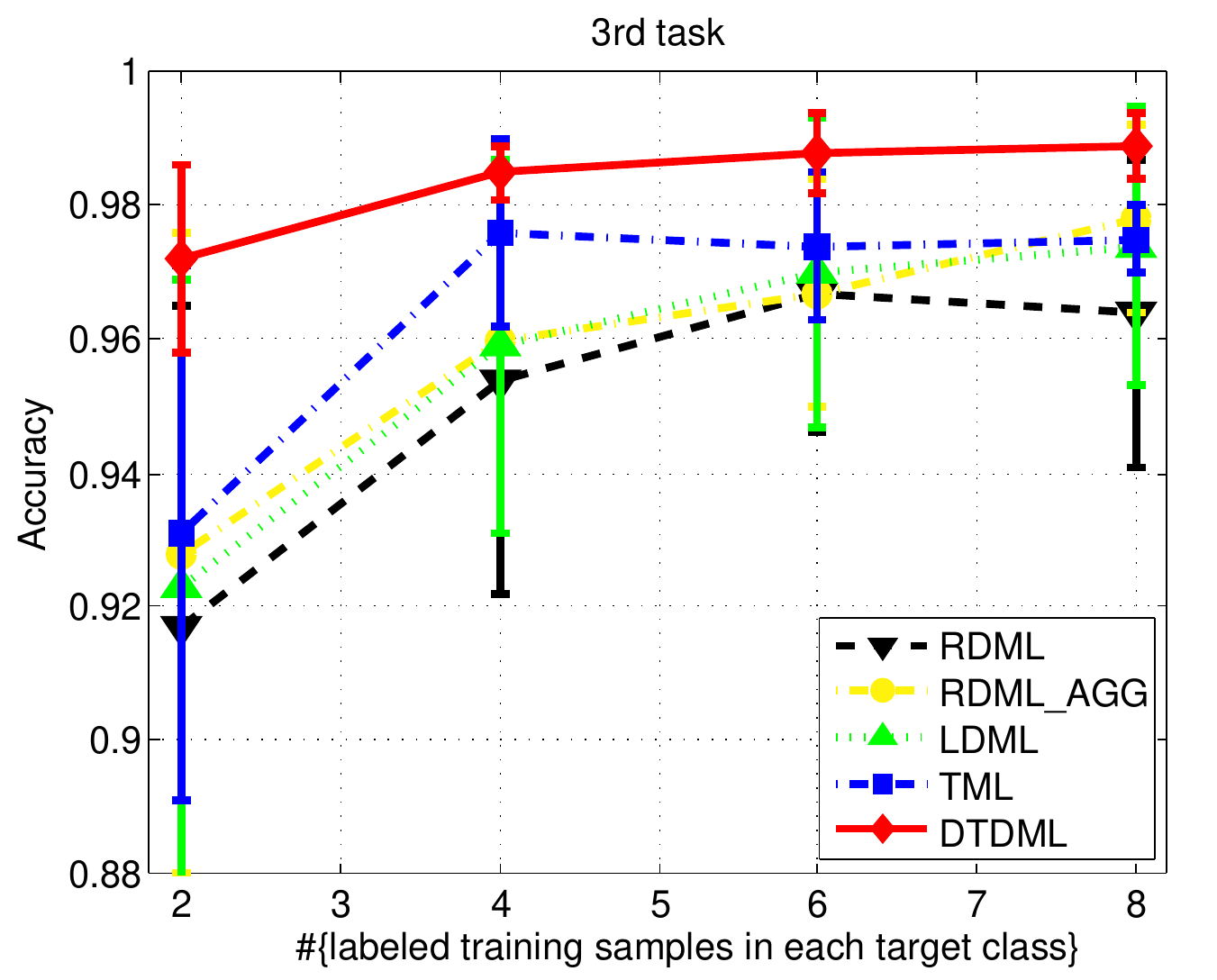}
}
\hfil
\subfigure{\includegraphics[width=0.65\columnwidth]{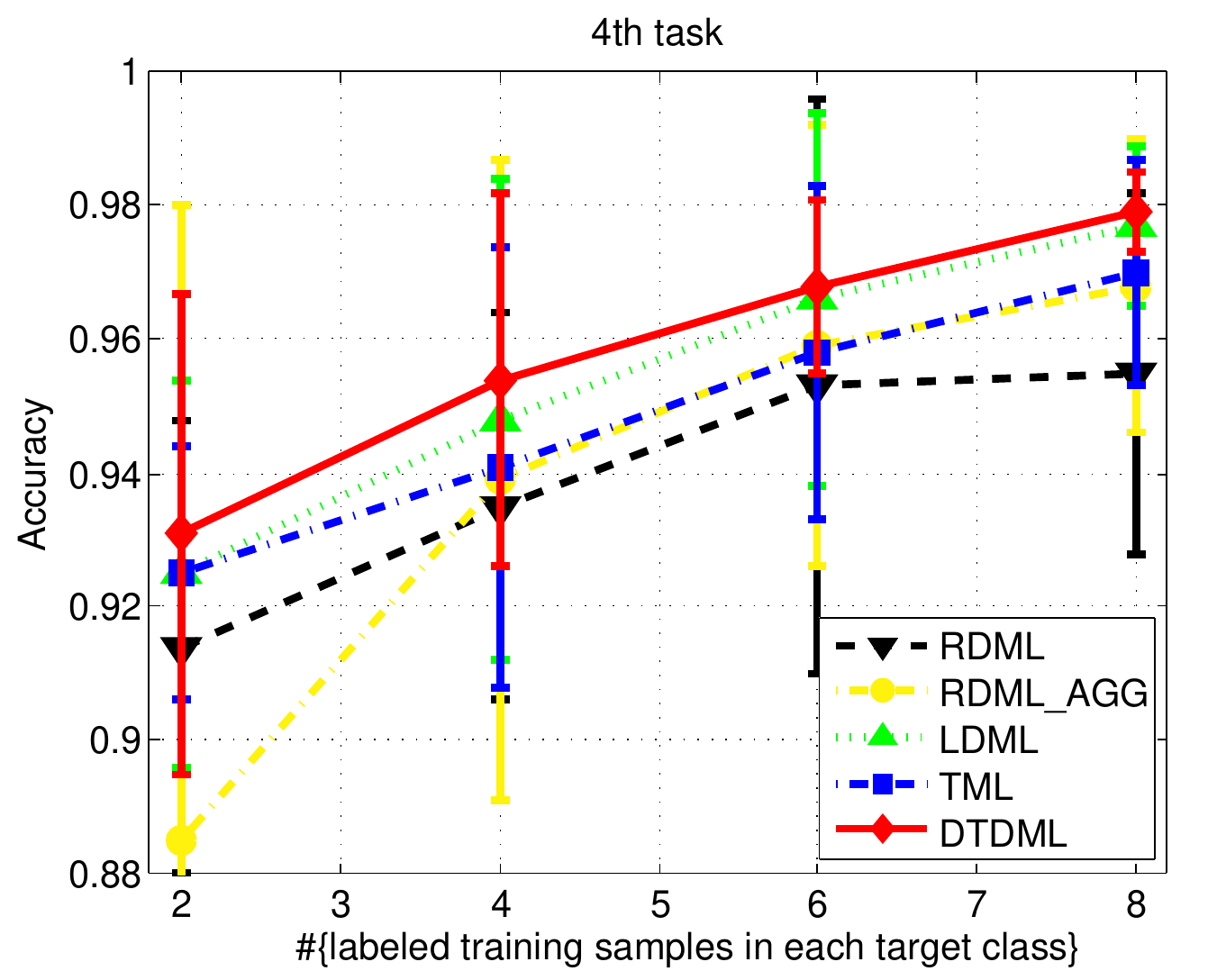}
}
\hfil
\subfigure{\includegraphics[width=0.65\columnwidth]{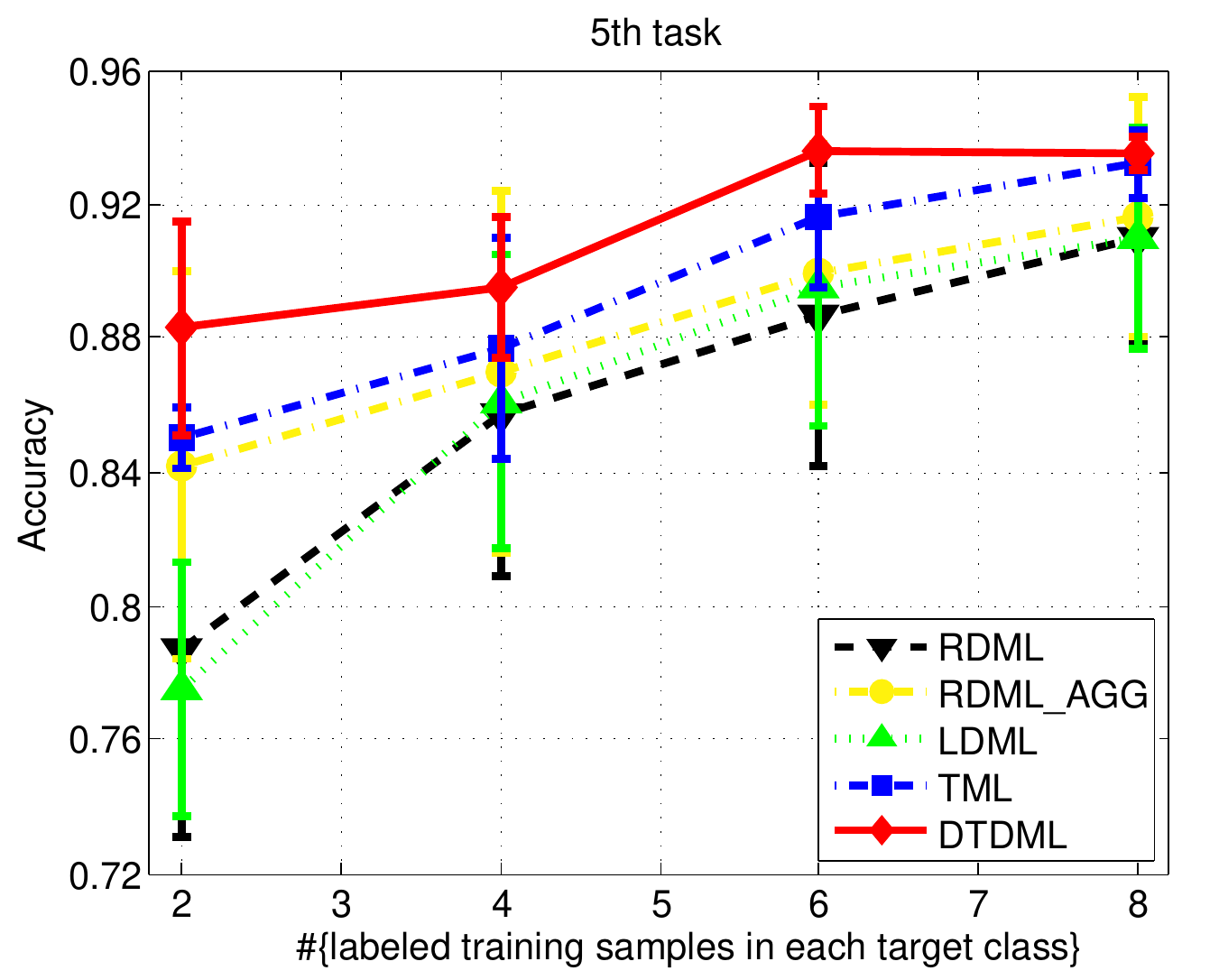}
}
\hfil
\subfigure{\includegraphics[width=0.65\columnwidth]{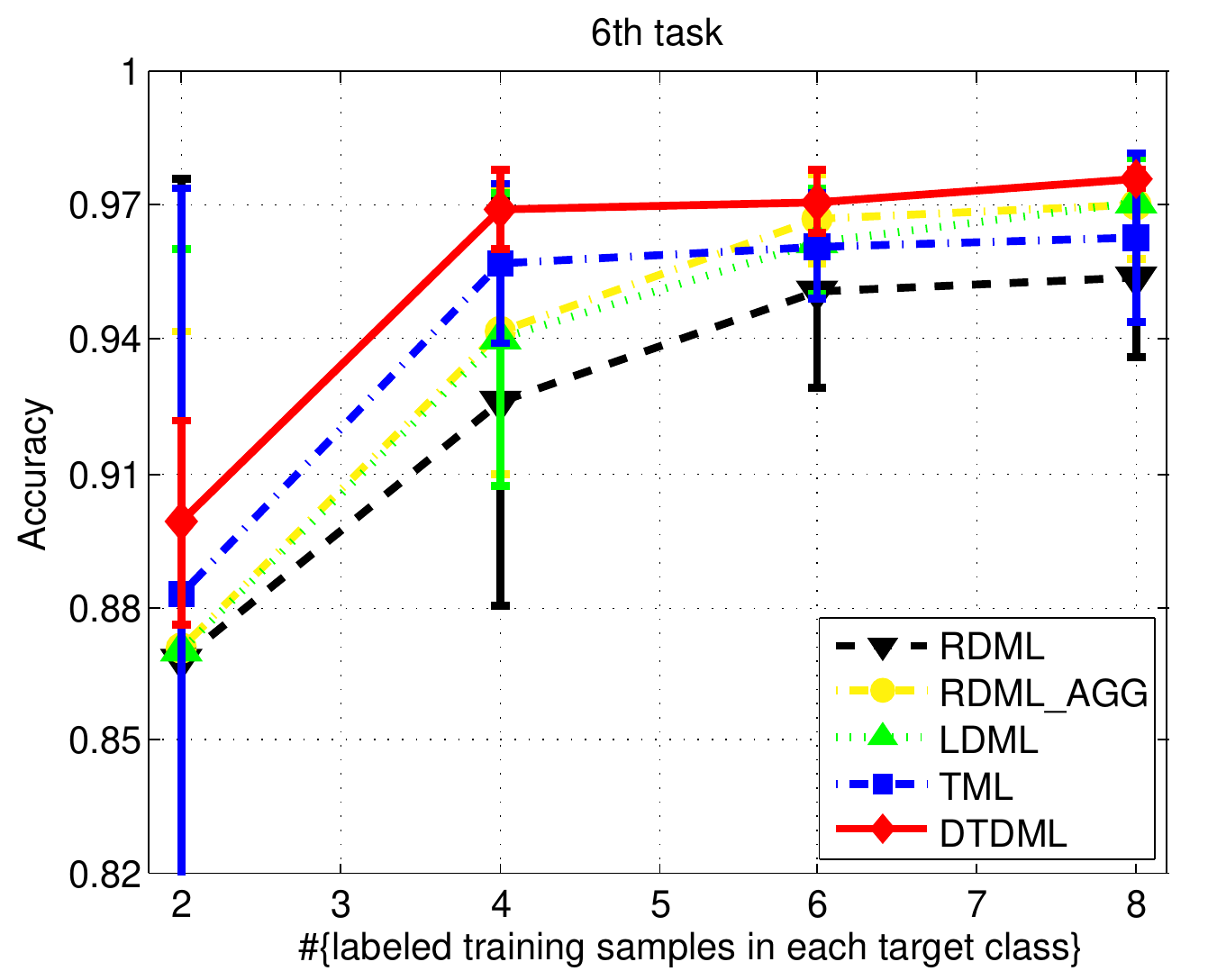}
}
\hfil
\subfigure{\includegraphics[width=0.65\columnwidth]{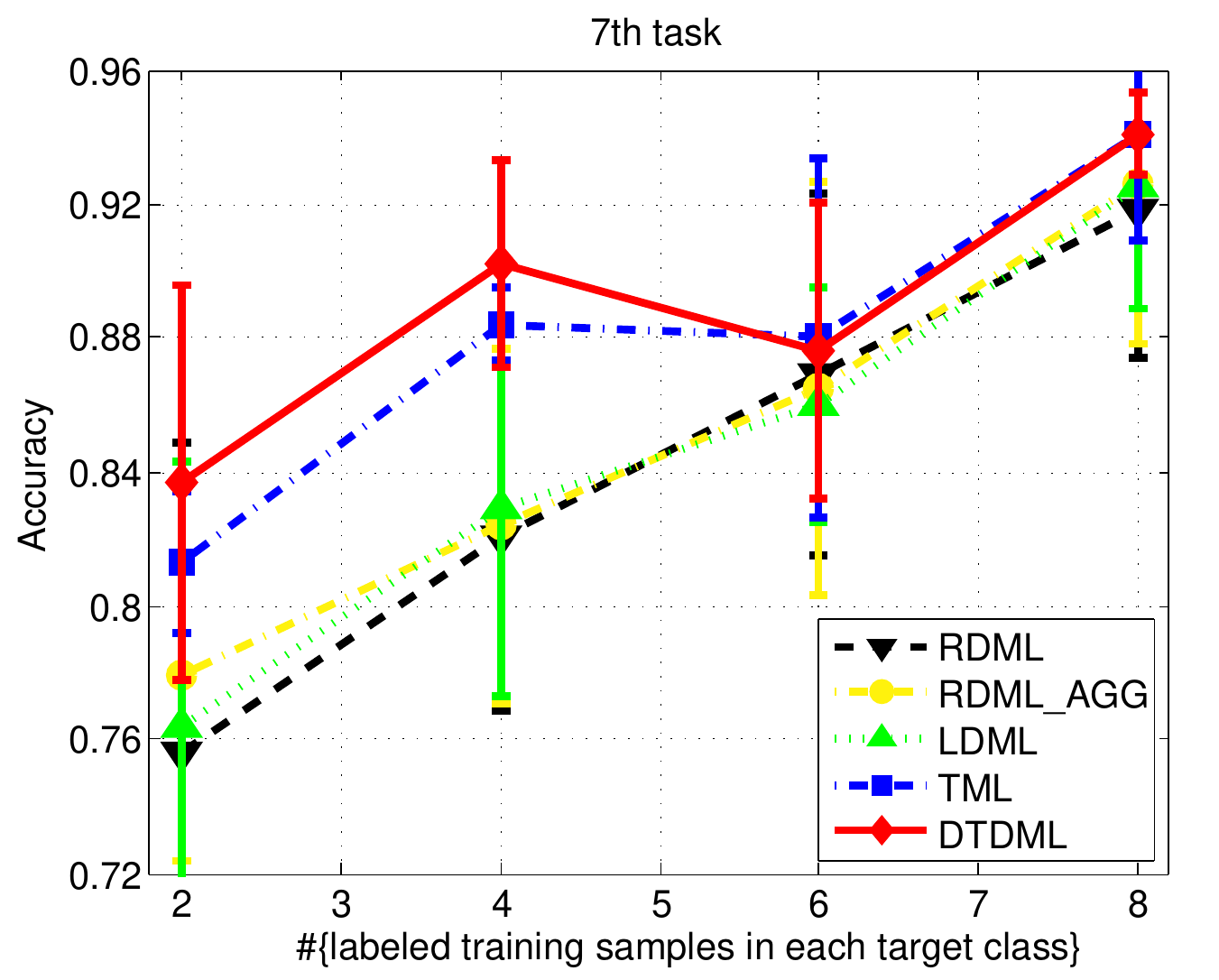}
}
\hfil
\subfigure{\includegraphics[width=0.65\columnwidth]{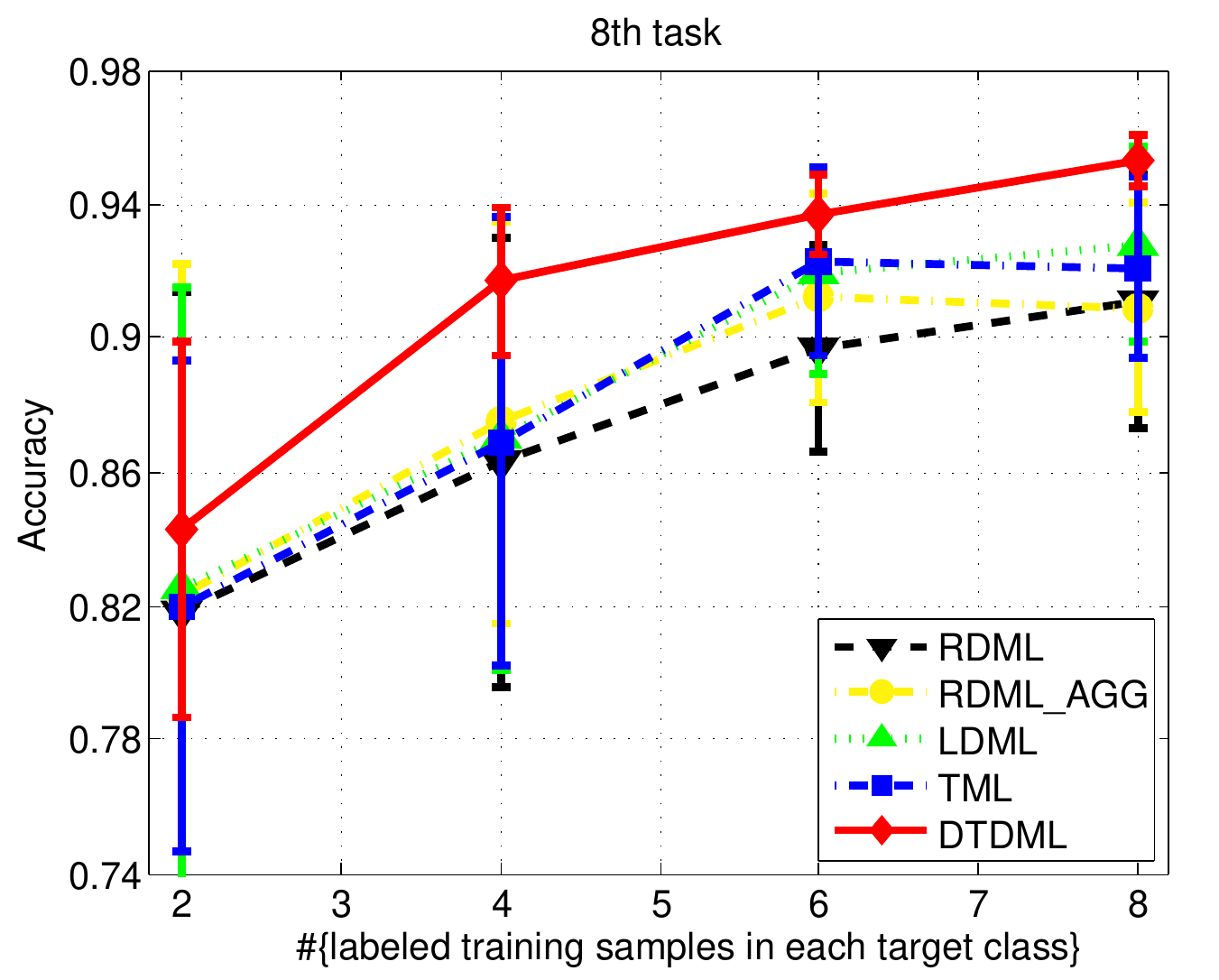}
}
\hfil
\subfigure{\includegraphics[width=0.65\columnwidth]{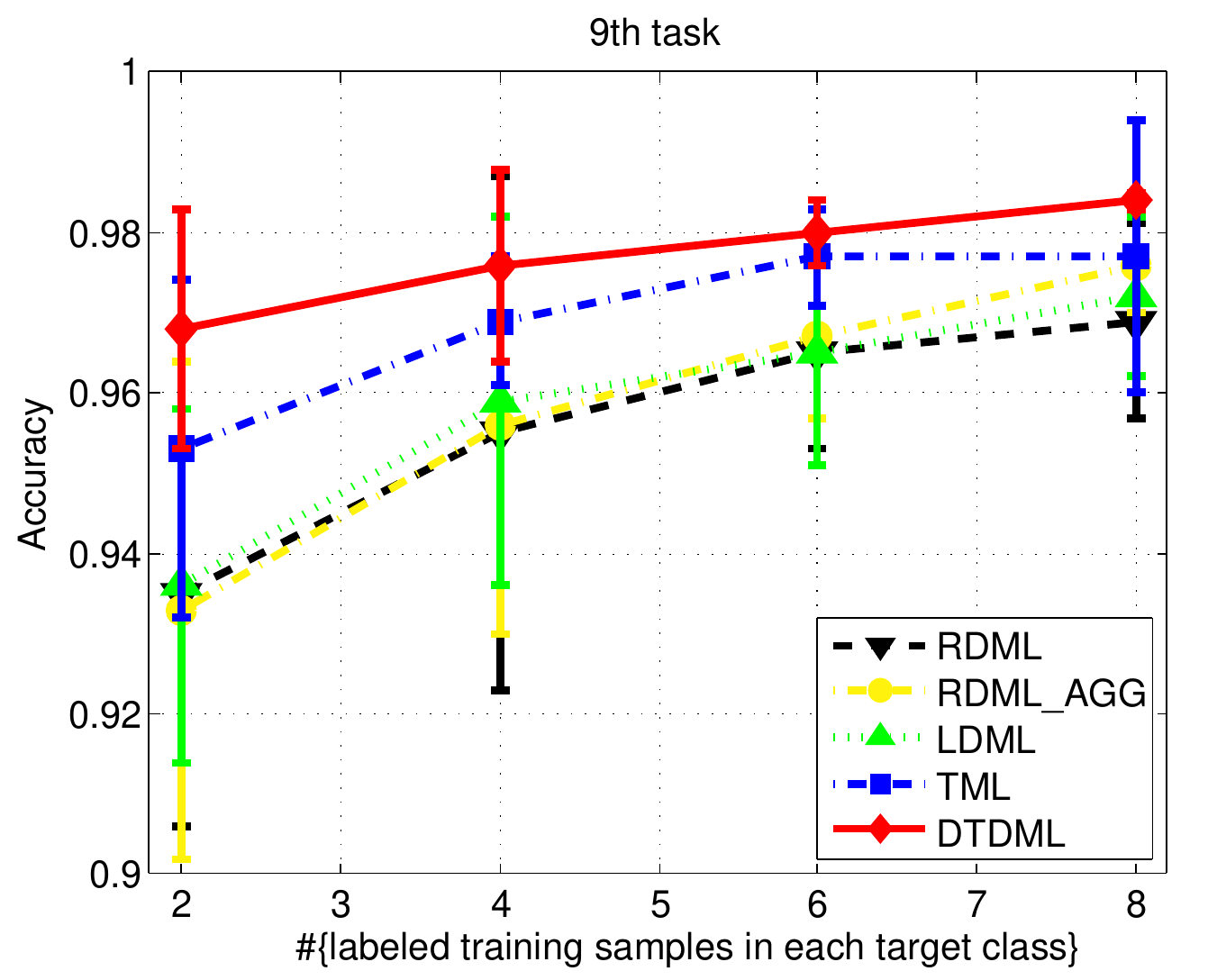}
}
\caption{Classification performance vs. the number of labeled training samples on the USPS digit dataset.}
\label{fig:Accuracy_Digit}
\end{figure*}

\subsubsection{A self-comparison of DTDML using source eigenvectors and random bases}
As depicted in this paper, the ``base metrics'' $B_r = u_r u_r^T$ that are utilized to construct the target metric $A = \sum_{r=1}^n \theta_r B_r$ can be derived from either the source eigenvectors, or other randomly generated bases. Therefore, we first investigate the performance of these two strategies, which are denoted as DTDML\_SE and DTDML\_RB, respectively. For DTDML\_SE, the $u_r, r=1,\ldots,n$ in problem (\ref{eq:DTDML_General_Formulation}) are eigenvectors of the source metrics, so the number of base metrics is fixed as $n = m \times d$. For DTDML\_RB, each $u_r$ is an eigenvector of some random matrix, and thus we can generate arbitrary number of base metrics. We randomly select one task from each of the two handwritten datasets, and report the results in Fig.~\ref{fig:Random_Bases}.

The results demonstrate that: 1) Even when the number of random bases $n_b$ is very small, e.g., $100$, we can still obtain satisfactory accuracy; 2) The accuracy of DTDML\_RB tends to be higher when $n_b$ is increased, but usually cannot outperform DTDML\_SE. To this end, we adopt DTDML\_SE in the following experiments. Another reason for choosing DTDML\_SE is to avoid tuning the additional parameter $n_b$.

\begin{figure*}
\centering
\subfigure{\includegraphics[width=0.65\columnwidth]{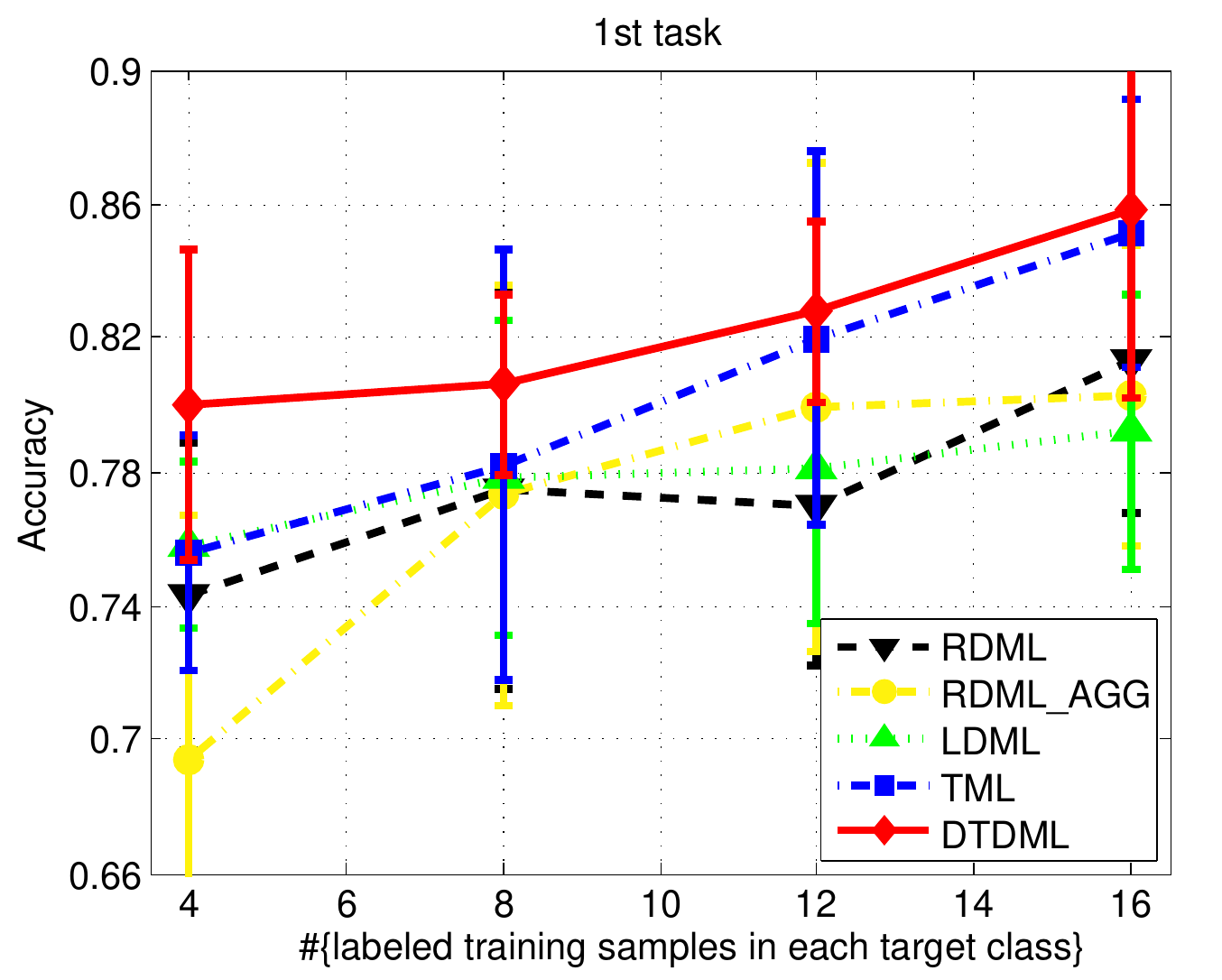}
}
\hfil
\subfigure{\includegraphics[width=0.65\columnwidth]{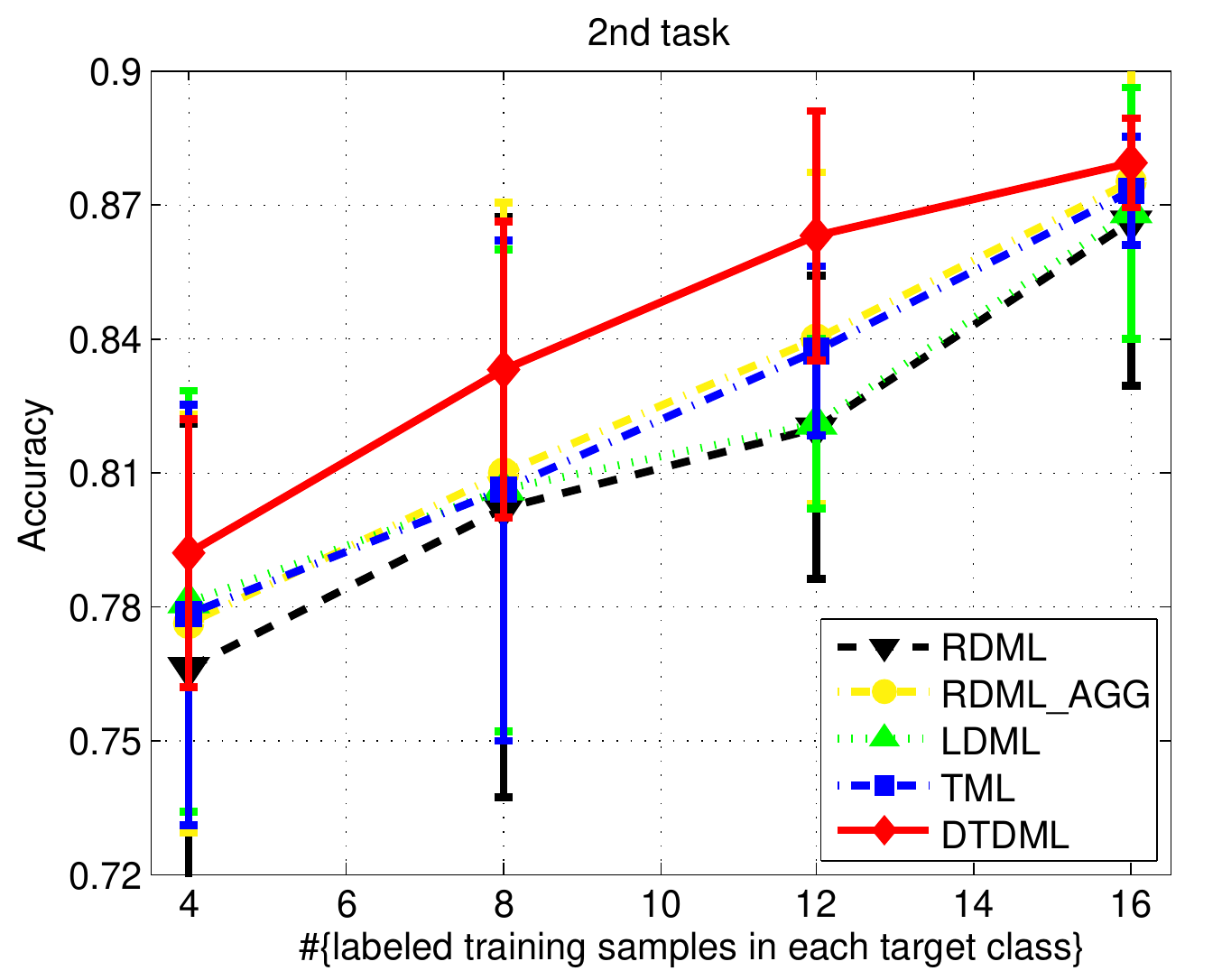}
}
\hfil
\subfigure{\includegraphics[width=0.65\columnwidth]{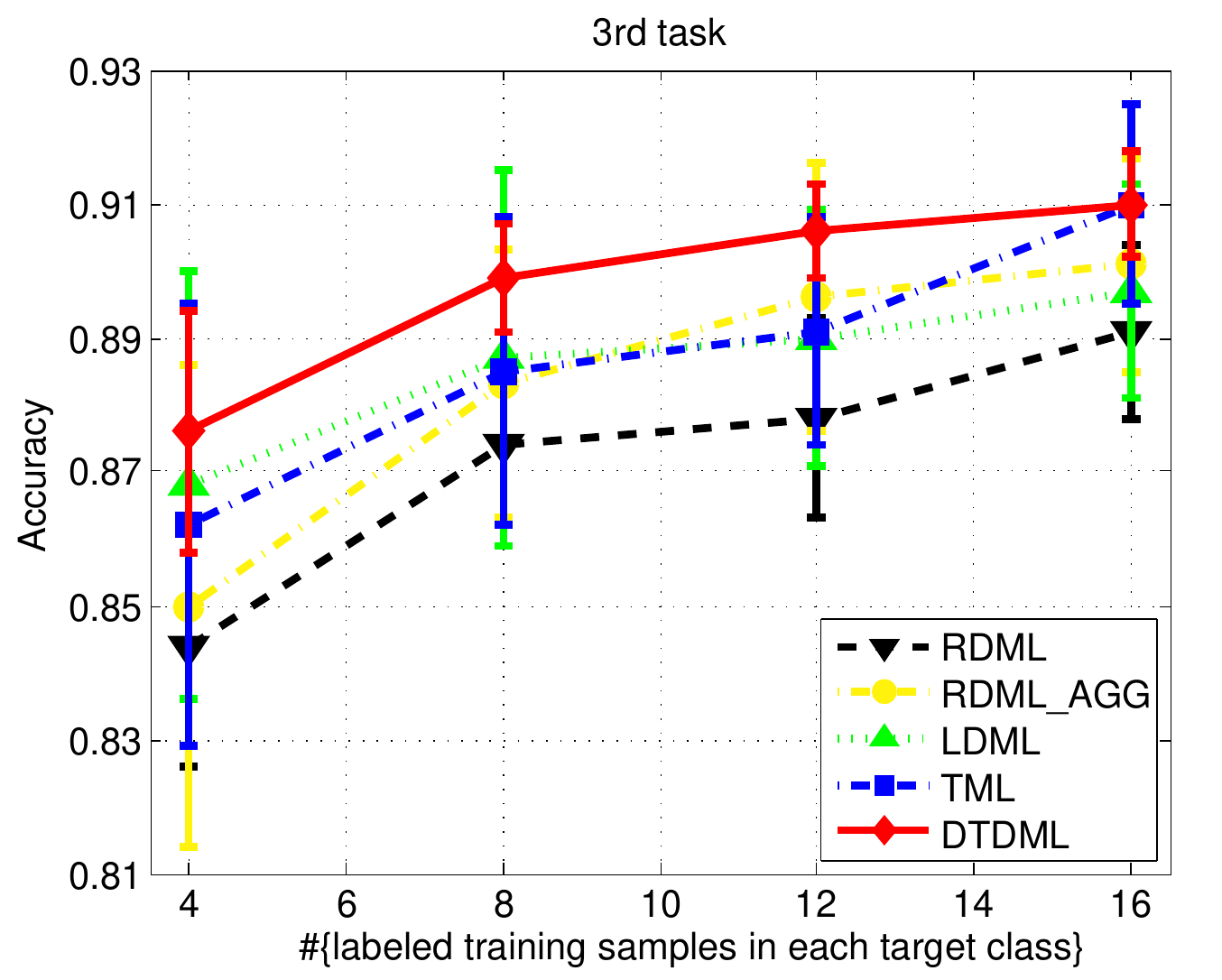}
}
\hfil
\subfigure{\includegraphics[width=0.65\columnwidth]{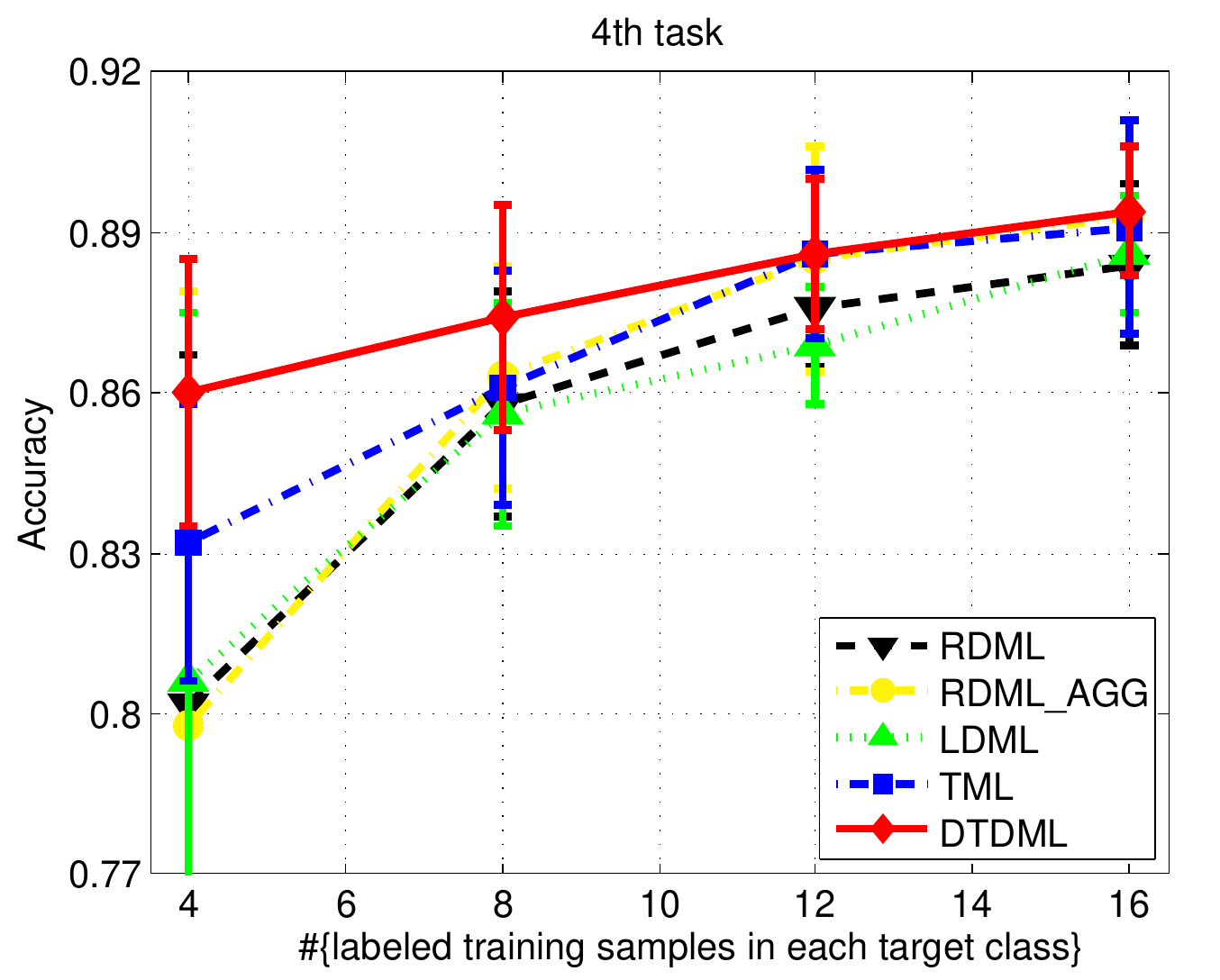}
}
\hfil
\subfigure{\includegraphics[width=0.65\columnwidth]{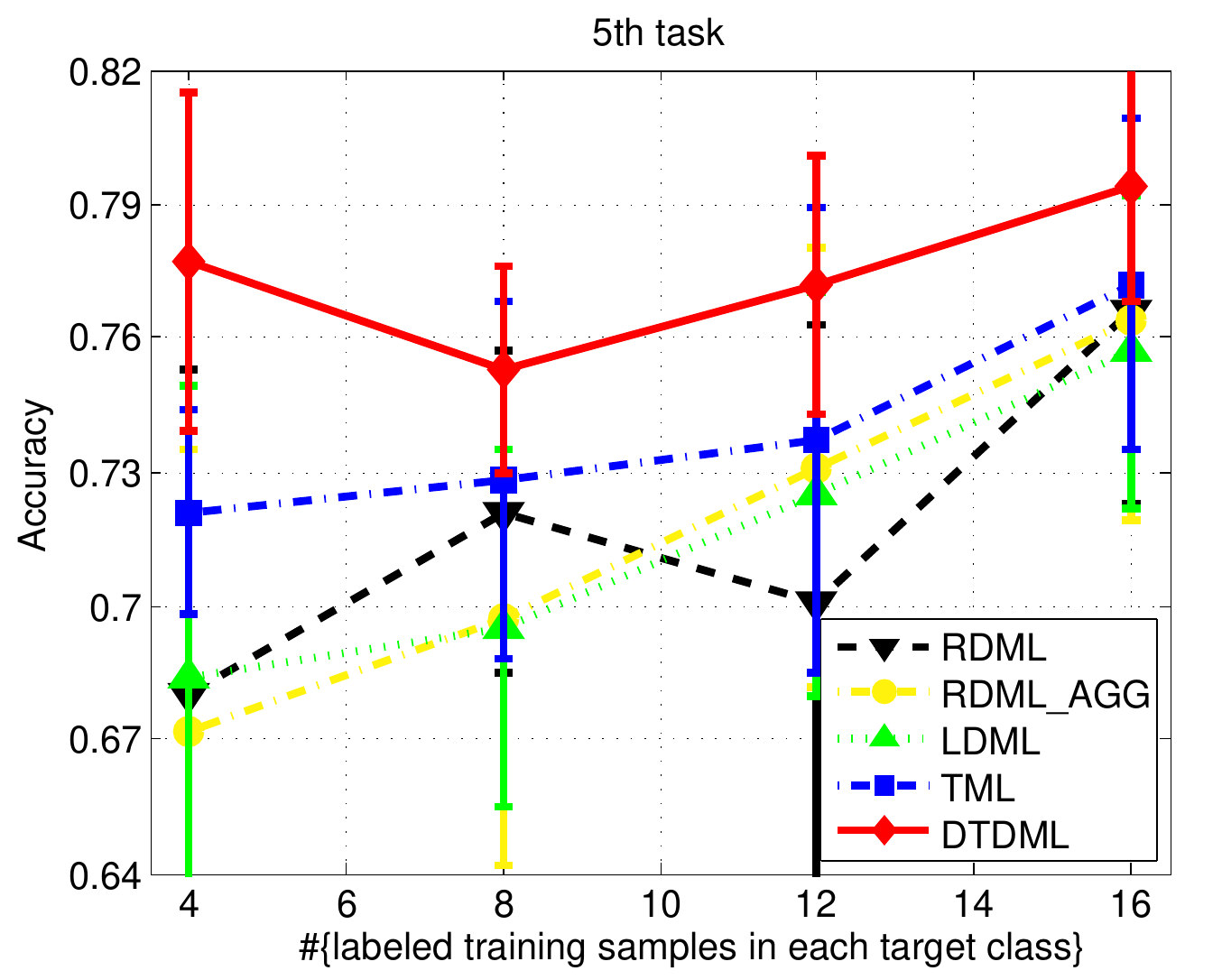}
}
\hfil
\subfigure{\includegraphics[width=0.65\columnwidth]{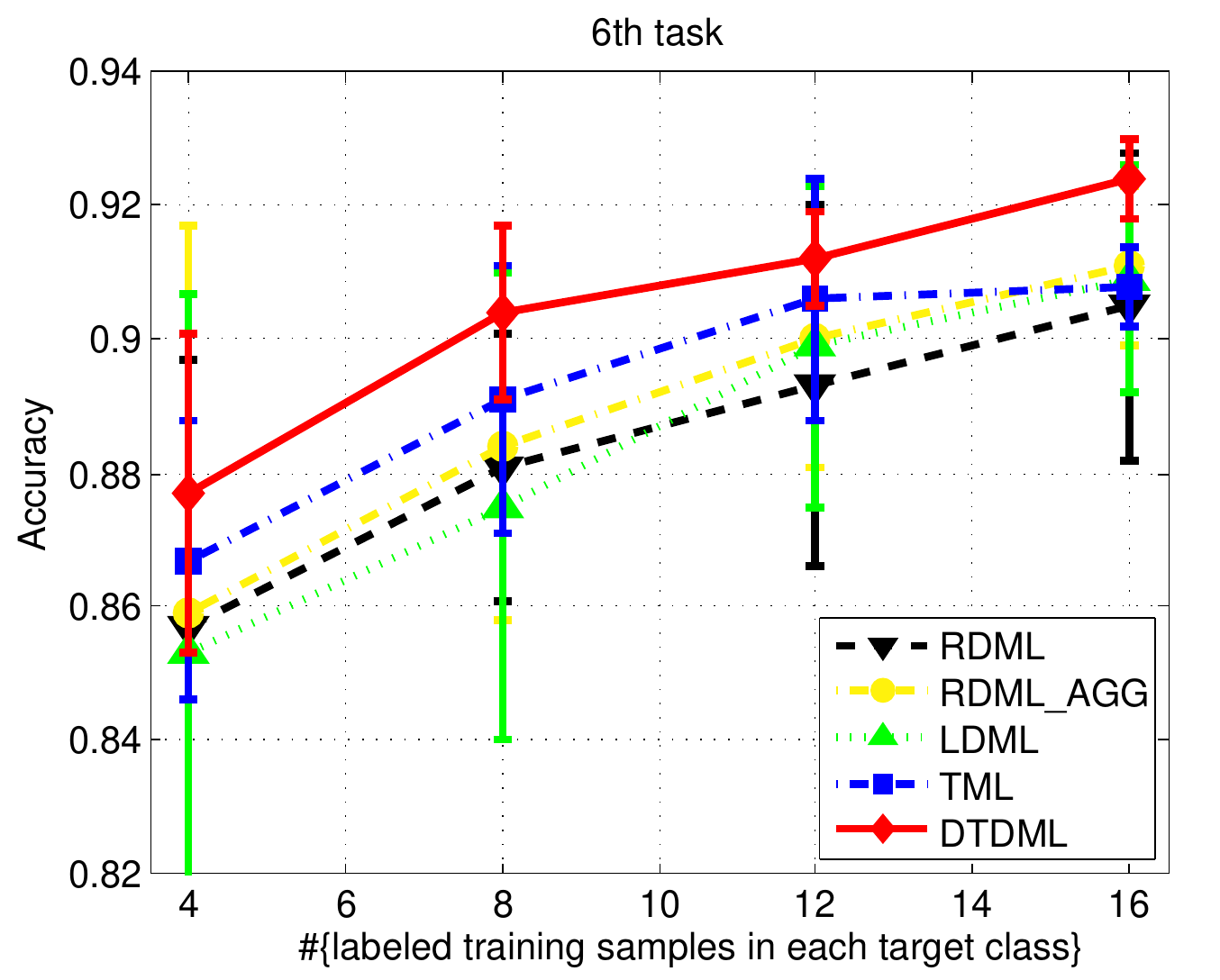}
}
\caption{Classification performance vs. the number of labeled training samples on the handwritten letter dataset.}
\label{fig:Accuracy_Letter}
\end{figure*}

\subsubsection{A comparison with the other algorithms}
The classification accuracies of different methods, under different settings on the digit dataset are shown in Fig.~\ref{fig:Accuracy_Digit}. We observe from the results that: 1) when the number of labeled training samples increases, the performance of all the compared methods tends to be better (higher mean accuracy and smaller variance); 2) the transfer metric learning algorithms (LDML, TML, DTDML) that utilize the source task information for target task learning are usually superior to RDML, which only learns on the target task. RDML is comparable to LDML on some tasks (e.g., the 5'th, 7'th, and 9'th task), which may be due to the finding of a bad local minima in LDML; 3) Overall, the performance of RDML\_AGG, which directly utilizes both the source and target training data without transfer, is better than RDML but worse than the transfer methods. This indicates that the distributions of the source and target datasets are different but related; 4) TML is better than LDML and RDML in most cases, while the proposed DTDML consistently outperforms all of them. In addition, we present the average performance over all settings in Table~\ref{tab:Average_Performance_USPS}. The results indicate a significant $3.2\%$ improvement compared with TML when using two labeled training samples. The level of improvement drops when more labeled samples are available. This is because DTDML has far fewer variables to learn than TML. The significance of this advantage gradually decreases since variable estimation can be steadily improved with an increase of labeled training samples. This indicates that the proposed algorithm is more suitable for the transfer scenario, since the labeled sample size of the target task is usually very small.

We report the performance on the letter dataset in Fig.~\ref{fig:Accuracy_Letter}. Similar to the digit classification, LDML is comparable to RDML and RDML\_AGG sometimes, and DTDML is superior to other methods significantly on almost all tasks. The average performance is presented in Table~\ref{tab:Average_Performance_Letter} and we observe a significant $3.8\%$ improvement compared against TML when using four labeled training samples.

\begin{table}[!t]
\setlength\tabcolsep{2pt}
\caption{Average performance over all tasks on USPS digit classification.}
\label{tab:Average_Performance_USPS}
\centering
\begin{tabular}{c||c|c|c|c}
\hline
Methods & 2 & 4 & 6 & 8 \\
\hline
RDML~\cite{RDML-R-Jin-et-al-NIPS-2009} & 0.855$\pm$0.071 & 0.903$\pm$0.046 & 0.926$\pm$0.035 & 0.942$\pm$0.029 \\
LDML~\cite{LDML-ZJ-Zha-et-al-IJCAI-2009} & 0.861$\pm$0.061 & 0.909$\pm$0.042 & 0.932$\pm$0.029 & 0.949$\pm$0.024 \\
TML~\cite{TML-Y-Zhang-and-DY-Yeung-TIST-2012} & 0.885$\pm$0.040 & 0.926$\pm$0.030 & 0.941$\pm$0.025 & 0.954$\pm$0.022 \\
DTDML & \textbf{0.913$\pm$0.031} & \textbf{0.943$\pm$0.018} & \textbf{0.955$\pm$0.014} & \textbf{0.968$\pm$0.006} \\
\hline
\end{tabular}
\end{table}

\begin{table}[!t]
\setlength\tabcolsep{2pt}
\caption{Average performance over all tasks on letter classification.}
\label{tab:Average_Performance_Letter}
\centering
\begin{tabular}{c||c|c|c|c}
\hline
Methods & 4 & 8 & 12 & 16 \\
\hline
RDML~\cite{RDML-R-Jin-et-al-NIPS-2009} & 0.782$\pm$0.049 & 0.818$\pm$0.036 & 0.823$\pm$0.033 & 0.854$\pm$0.029 \\
LDML~\cite{LDML-ZJ-Zha-et-al-IJCAI-2009} & 0.792$\pm$0.049 & 0.816$\pm$0.037 & 0.831$\pm$0.027 & 0.851$\pm$0.025 \\
TML~\cite{TML-Y-Zhang-and-DY-Yeung-TIST-2012} & 0.803$\pm$0.031 & 0.826$\pm$0.038 & 0.846$\pm$0.030 & 0.867$\pm$0.022 \\
DTDML & \textbf{0.835$\pm$0.042} & \textbf{0.845$\pm$0.021} & \textbf{0.857$\pm$0.019} & \textbf{0.877$\pm$0.020} \\
\hline
\end{tabular}
\end{table}

\subsection{Web image annotation}
This section provides details of the experiments conducted on a natural image dataset NUS-WIDE~\cite{NUSWIDE-TS-Chua-et-al-CIVR-2009} to further verify the effectiveness of the proposed algorithm. This dataset contains $269,648$ images and the features used in our experiments are $500$-D bag of visual words based on SIFT~\cite{DG-Lowe-IJCV-2004} descriptors. To perform a meaningful transfer, we select $12$ animal concepts: bear, bird, cat, cow, dog, elk, fish, fox, horse, tiger, whale, and zebra. For each concept, $100$ samples were randomly selected from the dataset.

In this set of experiments, the source task requires annotation of six randomly selected concepts, and the target task requires annotation of all others. Both are multi-class problems, but there is no difference in training compared to the binary case since the sample pairs are used and only the pair labels are needed. A pair of samples is labeled as positive if they are from the same class, and negative otherwise.

\begin{figure*}
\centering
\subfigure{\includegraphics[width=0.65\columnwidth]{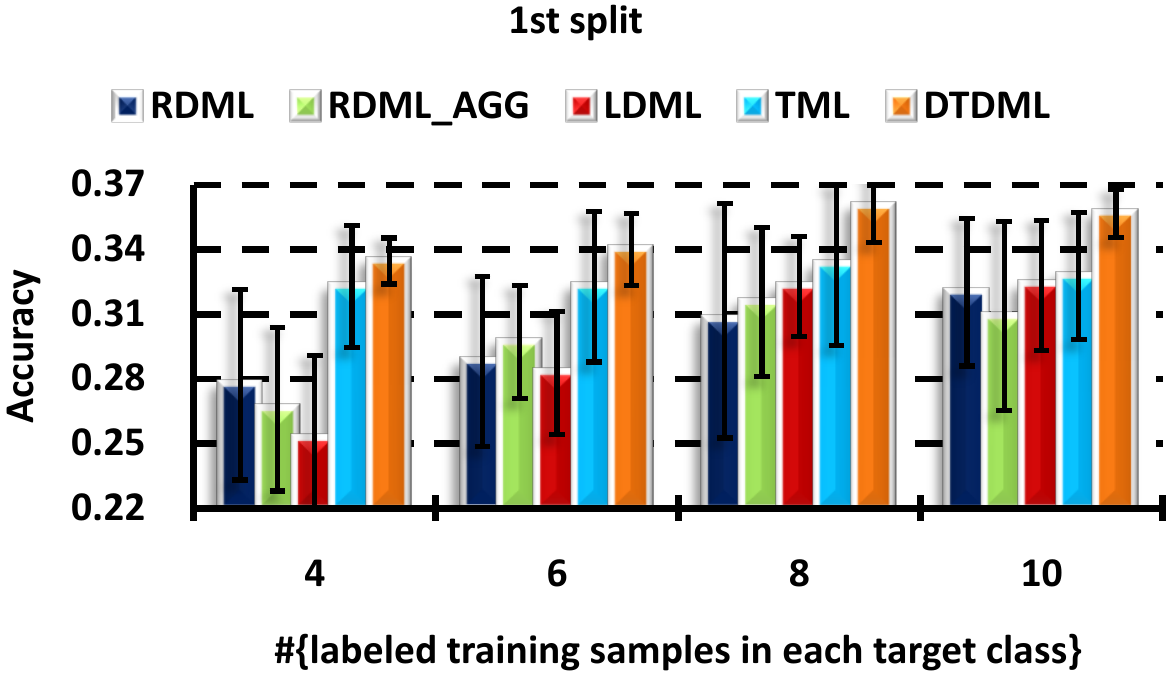}
}
\hfil
\subfigure{\includegraphics[width=0.65\columnwidth]{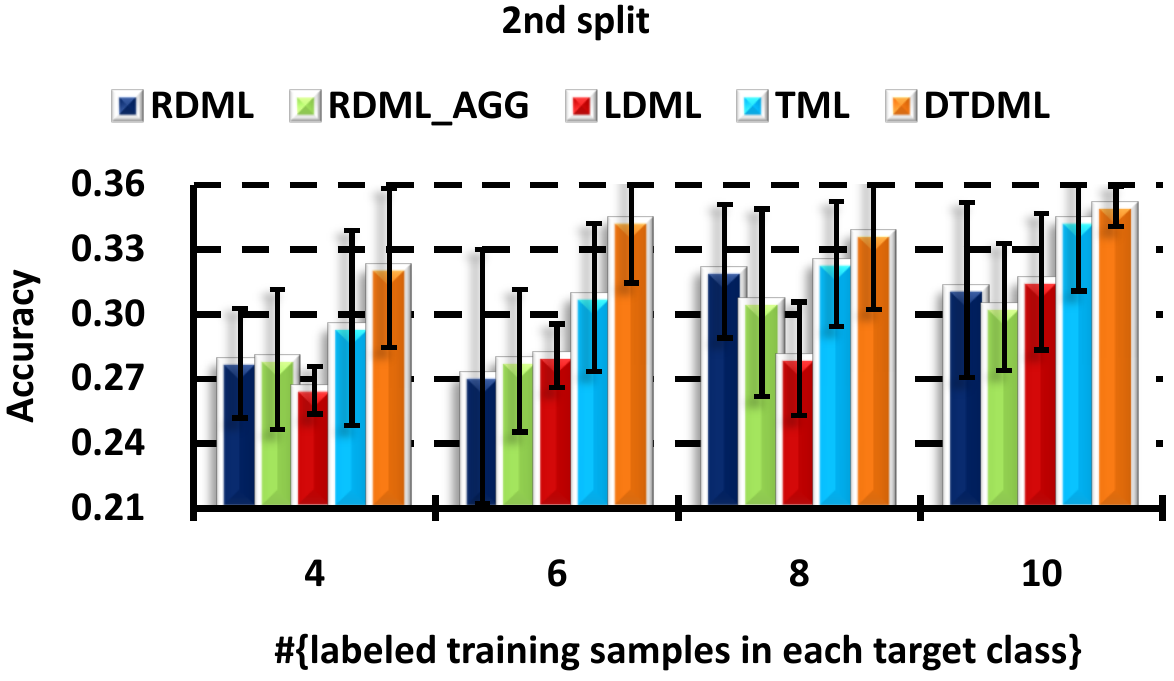}
}
\hfil
\subfigure{\includegraphics[width=0.65\columnwidth]{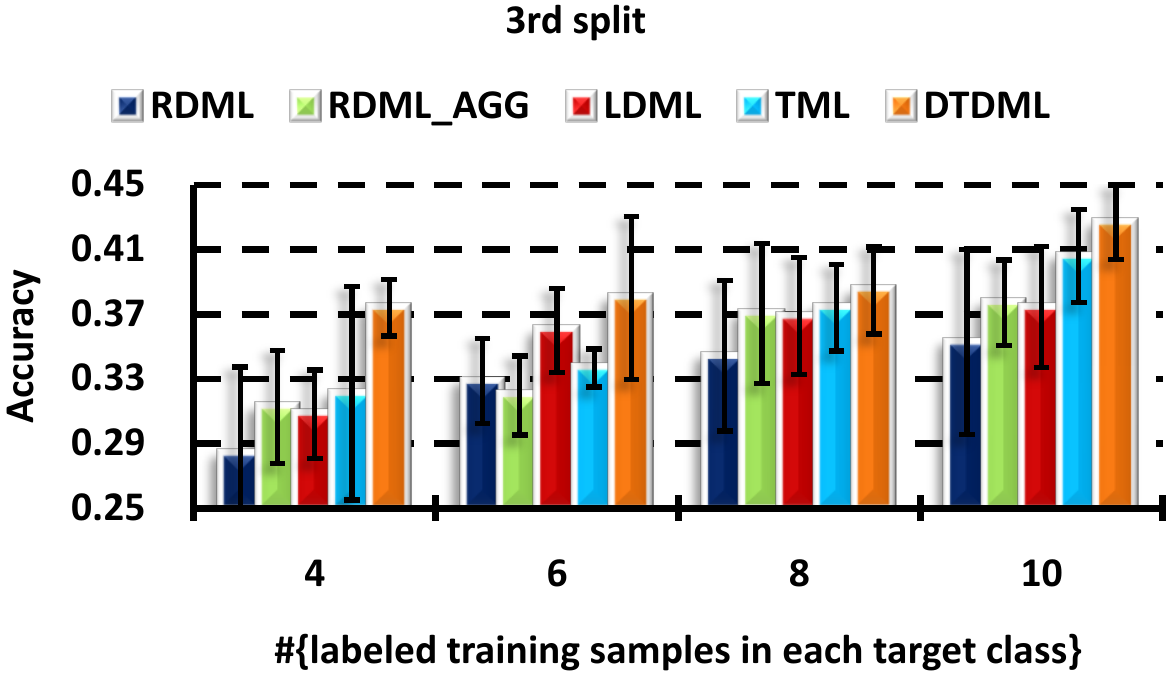}
}
\hfil
\subfigure{\includegraphics[width=0.65\columnwidth]{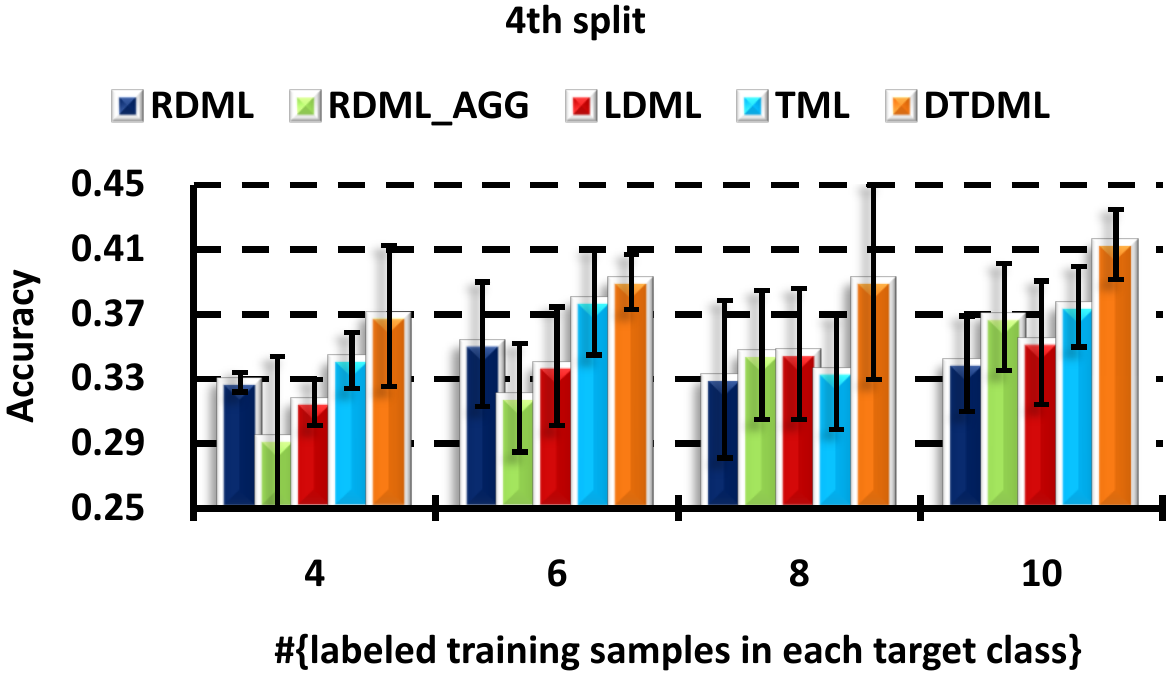}
}
\hfil
\subfigure{\includegraphics[width=0.65\columnwidth]{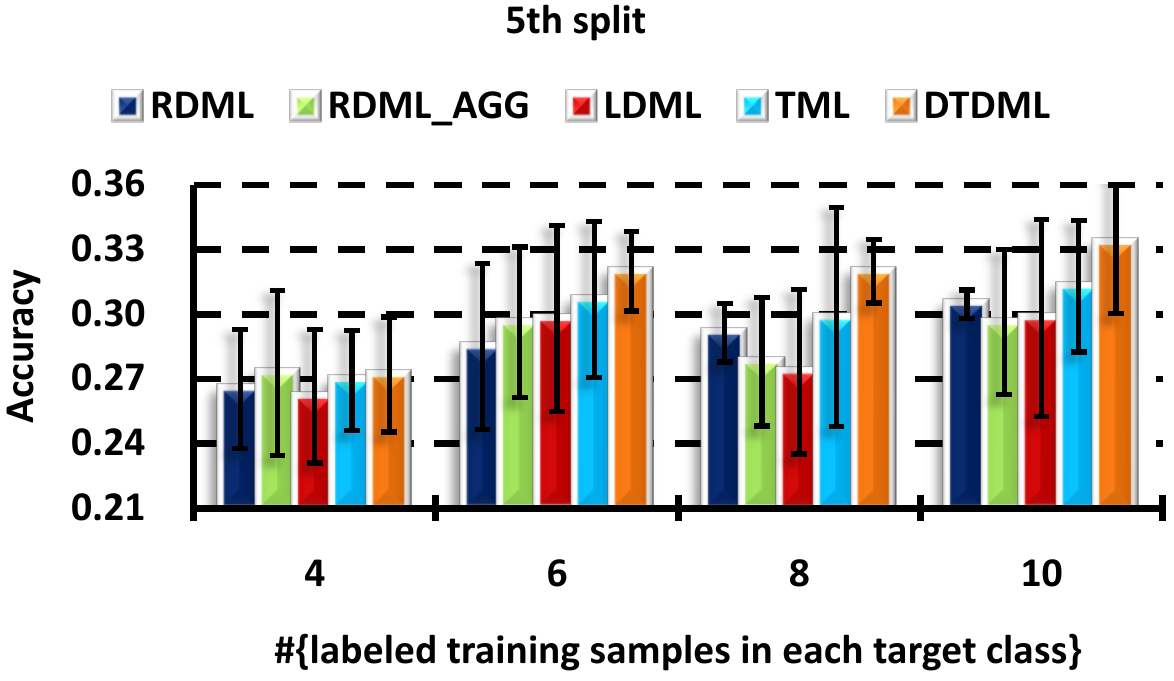}
}
\hfil
\subfigure{\includegraphics[width=0.65\columnwidth]{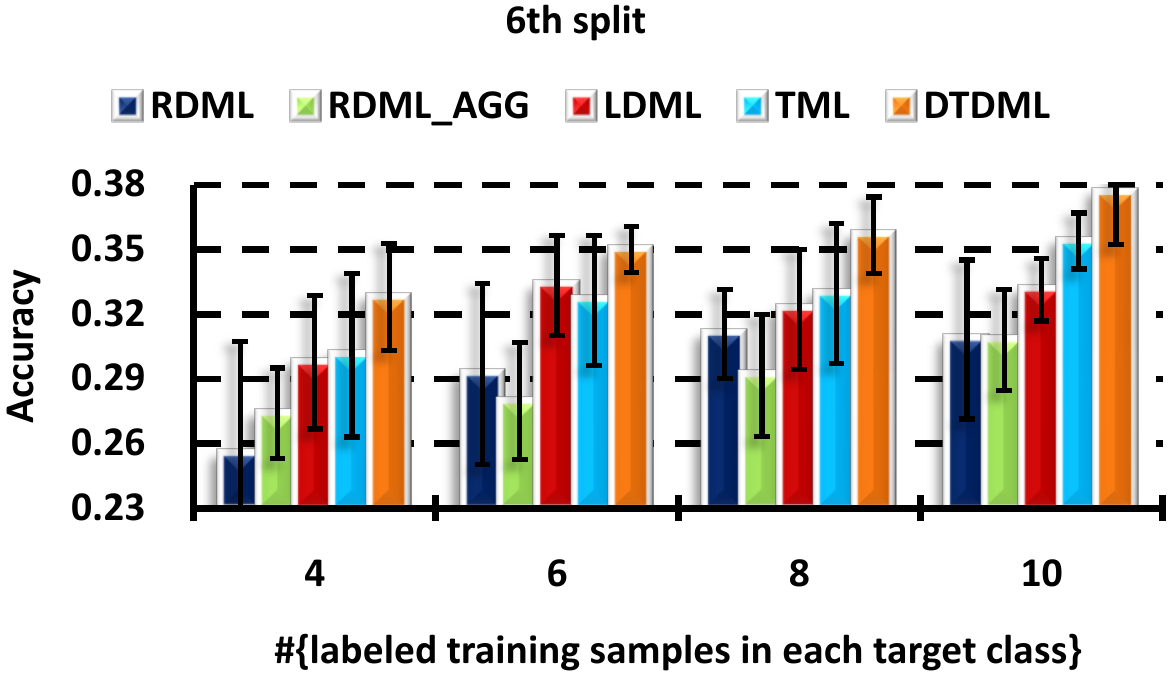}
}
\caption{Annotation performance vs. the number of labeled training samples on the NUS-WIDE dataset.}
\label{fig:Accuracy_NUS}
\end{figure*}

We perform six random splits of the concept set, and show the result of each split in Fig.~\ref{fig:Accuracy_NUS}. Similar conclusions can be obtained as in the handwritten image classification. DTDML always performs the best for all splits and in particular, we obtain an $8.1\%$ improvement on the average performance over all splits compared with TML when using four labeled samples.

\section{Conclusion and Discussion}
\label{sec:Conclusion}
Existing transfer metric learning approaches usually learn entries of the target metric directly, so the amount of variables is large, especially for the high dimensional image features. To resolve this problem, we have presented a decomposition based method called DTDML that assumes the target metric can be represented as a combination of ``base metrics''. DTDML has far less variables because we only have to learn the combination coefficients of the ``base metrics'', so better solutions can be obtained. In addition, we adopt Nesterov's optimal method to learn the coefficients and the optimization is quite efficient.

From the experimental validation on the popular handwritten image datasets and a challenging natural image dataset, we conclude that: 1) both source eigenvectors and random bases can be used to construct the target metric and the former performs a little better; 2) In the transfer scenario, using ``base metrics'' to induce the target metric is more effective than learning the target metric variables directly, even when the ``base metrics'' are randomly generated.

\appendices

\section{Proof of Theorem \ref{thm:Gradient_Lipschitz_g}}
\begin{IEEEproof}
\label{pf:Solution_MRMTL_LS_Linear}
According to (\ref{eq:Form_v}) and (\ref{eq:PieceWise_g}), we can calculate the gradient of $g_\sigma$ for the $k$'th sample as
\begin{numcases}
{\frac{\partial g_\sigma}{\partial \theta}=}{}\!\!\!
\begin{split}
\label{eq:Gradient_g_Derive}
0, \ \ \ \ \ \ \ \ \ \ \ \ \ \ \ \ \ & \ v_k = 0; \\
y_k h_k, \ \ \ \ \ \ \ \ \ \ \ \ \ \ & \ v_k = 1; \\
\frac{y_k h_k (-y_k (1 - \theta^T h_k))}{\sigma \|h_k\|_\infty}, \ & \ v_k = \frac{-y_k (1 - \theta^T h_k)}{\sigma \|h_k\|_\infty}.
\end{split}
\end{numcases}
This leads to (\ref{eq:Gradient_g}). Given function $g(x)$, for any $x^1$ and $x^2$, the Lipschitz constant $L$ satisfies
\begin{equation}
\label{eq:Lipschitz_g_Define}
\|\nabla g(x^1) - \nabla g(x^2)\|_2 \leq L \|x^1 - x^2\|_2.
\end{equation}
Hence the Lipschitz constant of $g_\sigma$ can be calculated from
\begin{equation}
\mathrm{max} \frac{\|\frac{\partial g_\sigma}{\partial \theta^1} - \frac{\partial g_\sigma}{\partial \theta^2}\|_2}{\|\theta^1 - \theta^2\|_2} \leq L^g.
\end{equation}
According to (\ref{eq:Gradient_g_Derive}), we have
\begin{numcases}
{\frac{\partial g_\sigma}{\partial \theta^1}\!-\!\frac{\partial g_\sigma}{\partial \theta^2}\!=\!}{}\!\!\!
\begin{split}
0, \ \ \ \ \ \ \ \ \ \ & y_k (1 - \theta^T h_k)\!<\!-\sigma \|h_k\|_\infty\ \mathrm{or} \\
& y_k (1\!-\!\theta^T h_k)\!>\!0; \\
\frac{h_k h_k^T (\theta^1\!-\!\theta^2)}{\sigma \|h_k\|_\infty}, & \mathrm{otherwise}.
\end{split}
\end{numcases}
Therefore,
\begin{equation}
\begin{split}
\mathop{\mathrm{max}}_k & \frac{\|h_k h_k^T (\theta^1 - \theta^2)\|_2}{\sigma \|h_k\|_\infty \|\theta^1 - \theta^2\|_2} \leq \mathop{\mathrm{max}}_k \frac{\|h_k h_k^T\|_2 \|(\theta^1 - \theta^2)\|_2}{\sigma \|h_k\|_\infty \|\theta^1 - \theta^2\|_2} \\
& = \frac{\|h_k h_k^T\|_2}{\sigma \|h_k\|_\infty} = L^g(h_k, y_k, \theta).
\end{split}
\end{equation}
To this end, the Lipschitz constant of $L^g(\theta)$ is calculated as
\begin{equation}
\begin{split}
& \sum_k L^g(h_k, y_k, \theta) \leq N' \mathop{\mathrm{max}}_k L^g(h_k, y_k, \theta) \\
& = \frac{N'}{\sigma} \mathop{\mathrm{max}}_k \frac{\|h_k h_k^T\|_2}{\|h_k\|_\infty} = L^g(\theta).
\end{split}
\end{equation}
This completes the proof.
\end{IEEEproof}

\section{Proof of Theorem \ref{thm:Uniform_Stability}}
\begin{IEEEproof}
\label{pf:Uniform_Stability}
Let's denote $F_\mathcal{N}(\theta) = P_\mathcal{N}(\theta) + Q(\theta)$, where $P_\mathcal{N}(\theta) = \frac{2}{N(N-1)} \sum_{i<j} V(A,z_i,z_j)$ and $Q(\theta) = \frac{\gamma_A}{2} \|A-A_S\|_F^2 + \gamma_C \|\theta\|_1$. It is obvious that both $P_\mathcal{N} (\theta)$ and $Q(\theta)$ are convex. We assume $\theta_\mathcal{N}$ and $\theta_{\mathcal{N}'}$ to be the minimizers of $F_\mathcal{N}(\theta)$ and $F_{\mathcal{N}'}(\theta)$, respectively, where $\mathcal{N}'$ is the collection of examples that replaces $z_i \in \mathcal{N}$ with another example $z_{i'}$.

Because the generalized Bregman divergence is non-negative and additive, we have
\begin{equation}
\label{eq:Bregman_Diver_FN_Than_Q}
\begin{split}
& B_{F_{\mathcal{N}}} (\theta_{\mathcal{N}'} \parallel \theta_{\mathcal{N}}) + B_{F_{\mathcal{N}'}} (\theta_{\mathcal{N}} \parallel \theta_{\mathcal{N}'}) \\
& \geq B_Q(\theta_{\mathcal{N}'} \parallel \theta_{\mathcal{N}}) + B_Q(\theta_{\mathcal{N}} \parallel \theta_{\mathcal{N}'}).
\end{split}
\end{equation}
Besides, $\partial Q(\theta_\mathcal{N}) / \partial \theta = \frac{\gamma_A}{2} \partial (\|A-A_S\|_F^2) / \partial \theta + \gamma_C \delta f(\theta)$, where $\delta f(\theta)$ is the subgradient of $\|\theta\|_1$, so we can obtain
\begin{equation}
\label{eq:Bregman_Diver_Q}
B_Q(\theta_{\mathcal{N}'} \parallel \theta_{\mathcal{N}}) + B_Q(\theta_{\mathcal{N}} \parallel \theta_{\mathcal{N}'}) = \gamma_A \|A_{\mathcal{N}'} - A_{\mathcal{N}}\|_F^2 + \gamma_C \Delta,
\end{equation}
where $\Delta = \|\theta_{\mathcal{N}}\|_1 - \langle \theta_{\mathcal{N}}, \mathrm{sgn}(\theta_{\mathcal{N}'}) \rangle + \|\theta_{\mathcal{N}'}\|_1 - \langle \theta_{\mathcal{N}'}, \mathrm{sgn}(\theta_{\mathcal{N}}) \rangle \geq 0$, $\mathrm{sgn}(\theta) = [\mathrm{sgn}(\theta_1), \ldots, \mathrm{sgn}(\theta_n)]^T$ and $\mathrm{sgn}(x) = 1$ if $x > 0$ and $-1$ otherwise.

According to (\ref{eq:Bregman_Diver_FN_Than_Q}) and (\ref{eq:Bregman_Diver_Q}), we have
\begin{equation}
\label{eq:Uniform_Stability_Derive}
\begin{split}
& \gamma_A \|A_{\mathcal{N}'} - A_{\mathcal{N}}\|_F^2 \leq B_{F_{\mathcal{N}}}(\theta_{\mathcal{N}'} \parallel \theta_{\mathcal{N}}) + B_{F_\mathcal{N}'}(\theta_{\mathcal{N}} \parallel \theta_{\mathcal{N}'}) \\
= & F_{\mathcal{N}}(\theta_{\mathcal{N}'}) - F_{\mathcal{N}}(\theta_{\mathcal{N}}) - \langle \theta_{\mathcal{N}'} - \theta_{\mathcal{N}}, \partial F_{\mathcal{N}}(\theta_{\mathcal{N}}) \rangle \\
& + F_{\mathcal{N}'}(\theta_{\mathcal{N}}) - F_{\mathcal{N}'}(\theta_{\mathcal{N}'}) - \langle \theta_{\mathcal{N}} - \theta_{\mathcal{N}'}, \partial F_{\mathcal{N}'}(\theta_{\mathcal{N}'}) \rangle \\
= & F_{\mathcal{N}}(\theta_{\mathcal{N}'}) - F_{\mathcal{N}}(\theta_{\mathcal{N}}) + F_{\mathcal{N}'}(\theta_{\mathcal{N}}) - F_{\mathcal{N}'}(\theta_{\mathcal{N}'}) \\
= & P_{\mathcal{N}}(\theta_{\mathcal{N}'}) - P_{\mathcal{N}}(\theta_{\mathcal{N}}) + P_{\mathcal{N}'}(\theta_{\mathcal{N}}) - P_{\mathcal{N}'}(\theta_{\mathcal{N}'}) \\
= & \frac{2}{N(N-1)} \left( \sum_{\mathcal{N}} V(A_{\mathcal{N}'},z_i,z_j) - \sum_{\mathcal{N}} V(A_{\mathcal{N}},z_i,z_j) \right. \\
& \left. + \sum_{\mathcal{N}'} V(A_{\mathcal{N}},z_{i'},z_j) - \sum_{\mathcal{N}'} V(A_{\mathcal{N}'},z_{i'},z_j) \right) \\
\leq & \frac{2}{N(N-1)} \left( \sum_{\mathcal{N}} |V(A_{\mathcal{N}'},z_i,z_j) - V(A_{\mathcal{N}},z_i,z_j)| \right. \\
& \left. + \sum_{\mathcal{N}'} |V(A_{\mathcal{N}},z_{i'},z_j) - V(A_{\mathcal{N}'},z_{i'},z_j)| \right) \\
\leq & \frac{16LR^2}{N} \|A_{\mathcal{N}'}-A_{\mathcal{N}}\|_F.
\end{split}
\end{equation}
The second equality holds because $\theta_\mathcal{N}$ and $\theta_{\mathcal{N}'}$ are minimizers of $F_\mathcal{N}(\theta)$ and $F_{\mathcal{N}'}(\theta)$ respectively, which implies that $\partial F_\mathcal{N}(\theta_\mathcal{N}) = \partial F_{\mathcal{N}'}(\theta_{\mathcal{N}'}) = 0$. The last inequality holds because of Lemma~\ref{lema:Metric_Diff_For_Bound}. By comparing the left and right side of (\ref{eq:Uniform_Stability_Derive}), we obtain
\begin{equation}
\|A_{\mathcal{N}}-A_{\mathcal{N}'}\|_F \leq \frac{16LR^2}{\gamma_A N}.
\end{equation}
By further utilizing Lemma~\ref{lema:Metric_Diff_For_Bound}, i.e., $|V(A_\mathcal{N},z_i,z_j) - V(A_{\mathcal{N}'},z_i,z_j)| \leq 4LR^2 \|A_\mathcal{N} - A_{\mathcal{N}'}\|_F$, we have
\begin{equation}
|V(A_\mathcal{N},z_i,z_j) - V(A_{\mathcal{N}'},z_i,z_j)| \leq \frac{64L^2R^4}{\gamma_A N}.
\end{equation}
This completes the proof.
\end{IEEEproof}

\section{Proof of Lemma \ref{lema:Inequalities_For_Bound}}
\begin{IEEEproof}
\label{pf:Inequalities_For_Bound}
Because $\theta_S$ is a solution of $A = A_S$, so we have
\begin{equation}
\begin{split}
& \frac{2}{N(N-1)} \sum_{i<j} V(A_\mathcal{N}, z_i, z_j) + \frac{\gamma_A}{2} \|A_\mathcal{N} - A_S\|_F^2 \\
& + \frac{\gamma_B}{2} \|\alpha\|_2^2 + \gamma_C \|\theta_\mathcal{N}\|_1 \\
\leq & \frac{2}{N(N-1)} \sum_{i<j} V(A_S, z_i, z_j) + \frac{\gamma_B}{2} \|\alpha\|_2^2 + \gamma_C \|\theta_S\|_1,
\end{split}
\end{equation}
This leads to
\begin{equation}
\begin{split}
\frac{\gamma_A}{2} \|A_\mathcal{N} - A_S\|_F^2 \leq & \frac{2}{N(N-1)} \sum_{i<j} V(A_S, z_i, z_j) \\
& + \gamma_C (\|\theta_S\|_1 - \|\theta_\mathcal{N}\|_1).
\end{split}
\end{equation}
since $\frac{2}{N(N-1)} \sum_{i<j} V(A_\mathcal{N},z_i,z_j) \geq 0$. Therefore, we have $\|A_\mathcal{N} - A_S\|_F \leq \sqrt{ \left( 2(g_{A_S} + \gamma_C(\|\theta_S\|_1 - \|\theta_\mathcal{N}\|_1)) \right) / \gamma_A }$. The same procedure can be applied to bounding $\|A_{\mathcal{N}'} - A_S\|_F$.
\end{IEEEproof}

\section{Proof of Theorem \ref{thm:Generalization_Error_Bound}}
\begin{IEEEproof}
\label{pf:Generalization_Error_Bound}
Let $\Phi(A_\mathcal{N}) = R(A_\mathcal{N}) - R_\mathcal{N}(A_\mathcal{N})$. It follows from~\cite{O-Bousquet-and-A-Elisseeff-JMLR-2002} that $\Phi(A_\mathcal{N}) \leq 2\beta$. Besides,
\begin{equation}
\begin{split}
& |\Phi(A_{\mathcal{N}}) - \Phi(A_{\mathcal{N}'})| \\
= & |R(A_{\mathcal{N}}) - R_{\mathcal{N}}(A_{\mathcal{N}}) - R(A_{\mathcal{N}'}) + R_{\mathcal{N}'}(A_{\mathcal{N}'})| \\
\leq & |R(A_{\mathcal{N}}) - R(A_{\mathcal{N}'})| + |R_{\mathcal{N}}(A_{\mathcal{N}}) - R_{\mathcal{N}'}(A_{\mathcal{N}'})| \\
\leq & \beta + \frac{2}{N(N-1)} \left( \frac{N(N-1)}{2} - (N-1) \right) \beta \\
& + \left| \frac{2}{N(N-1)} \left( \sum_{j\neq i} V(A_{\mathcal{N}},z_i,z_j) - \sum_{j\neq {i'}} V(A_{\mathcal{N}'},z_{i'},z_j) \right) \right| \\
\leq & 2\beta + \frac{2}{N(N-1)} \sum_{j\neq i} |V(A_{\mathcal{N}},z_i,z_j) - V(A_S,z_i,z_j)| \\
& + \frac{2}{N(N-1)} \sum_{j\neq {i'}} |V(A_{\mathcal{N}'},z_{i'},z_j) - V(A_S,z_{i'},z_j)| + \frac{4 g_{A_S}}{N} \\
\leq & 2\beta + \frac{4 g_{A_S}}{N} + \frac{8LR^2(\|A_{\mathcal{N}} - A_S\|_F + \|A_{\mathcal{N}'} - A_S\|_F)}{N} \\
\leq & 2\beta + \frac{4 g_{A_S}}{N} + \Big[ \left( 8\sqrt{2}LR^2 (\sqrt{g_{A_S} + \gamma_C (\|\theta_S\|_1 - \|\theta_{\mathcal{N}}\|_1)} \right. \\
& \left. + \sqrt{g_{A_S} + \gamma_C \|\theta_S\|_1}) \right) / (\sqrt{\gamma_A} N) \Big] \triangleq M'.
\end{split}
\end{equation}
where $g_{A_S}=\sup_{z_i,z_j} V(A_S,z_i,z_j)$ is the largest loss when the distance metric is $A_S$. The last inequality holds because of Lemma~\ref{lema:Inequalities_For_Bound}.

Given $\delta > 0$, using the McDiarmid inequality, with probability at least $1 - \delta$, we have
\begin{equation}
\Phi(A_{\mathcal{N}}) \leq E[\Phi(A_{\mathcal{N}})] + N M' \sqrt{\frac{\ln(1/\delta)}{2N}}.
\end{equation}
In addition, we can conclude that $E[\Phi(A_\mathcal{N})] \leq 2\beta$:
\begin{equation}
\begin{split}
& |E\Phi(A_{\mathcal{N}})| \\
= & |E\left( R(A_{\mathcal{N}}) - R(A_{\mathcal{N}'}) + R(A_{\mathcal{N}'}) - R_{\mathcal{N}}(A_{\mathcal{N}}) \right)| \\
\leq & |E(R(A_{\mathcal{N}}) - R(A_{\mathcal{N}'}))| + |E(R(A_{\mathcal{N}'}) - R_{\mathcal{N}}(A_{\mathcal{N}}))| \\
= & 2|E(V(A_{\mathcal{N}},z_i,z_j) - V(A_{\mathcal{N}'},z_i,z_j))| \leq 2\beta.
\end{split}
\end{equation}
This completes the proof.
\end{IEEEproof}


%\section{}

%\section*{Acknowledgment}

\ifCLASSOPTIONcaptionsoff
  \newpage
\fi



\bibliographystyle{IEEEtran}
\bibliography{./TIP-11260-2013}
\end{document}